\documentclass[10pt,twocolumn,letterpaper]{article}

\usepackage[pagenumbers]{cvpr} 

\usepackage{graphicx}
\usepackage{subcaption}
\usepackage{pifont}

\usepackage{booktabs}
\usepackage{multirow}

\usepackage{algorithm}
\usepackage{algorithmic}

\usepackage[dvipsnames]{xcolor}
\usepackage[table,xcdraw]{xcolor}

\usepackage[normalem]{ulem}

\definecolor{cvprblue}{rgb}{0.21,0.49,0.74}
\usepackage[pagebackref,breaklinks,colorlinks,allcolors=cvprblue]{hyperref}
\def\paperID{5892} 
\def\confName{CVPR}
\def\confYear{2026}

\title{\includegraphics[width=0.04\textwidth]{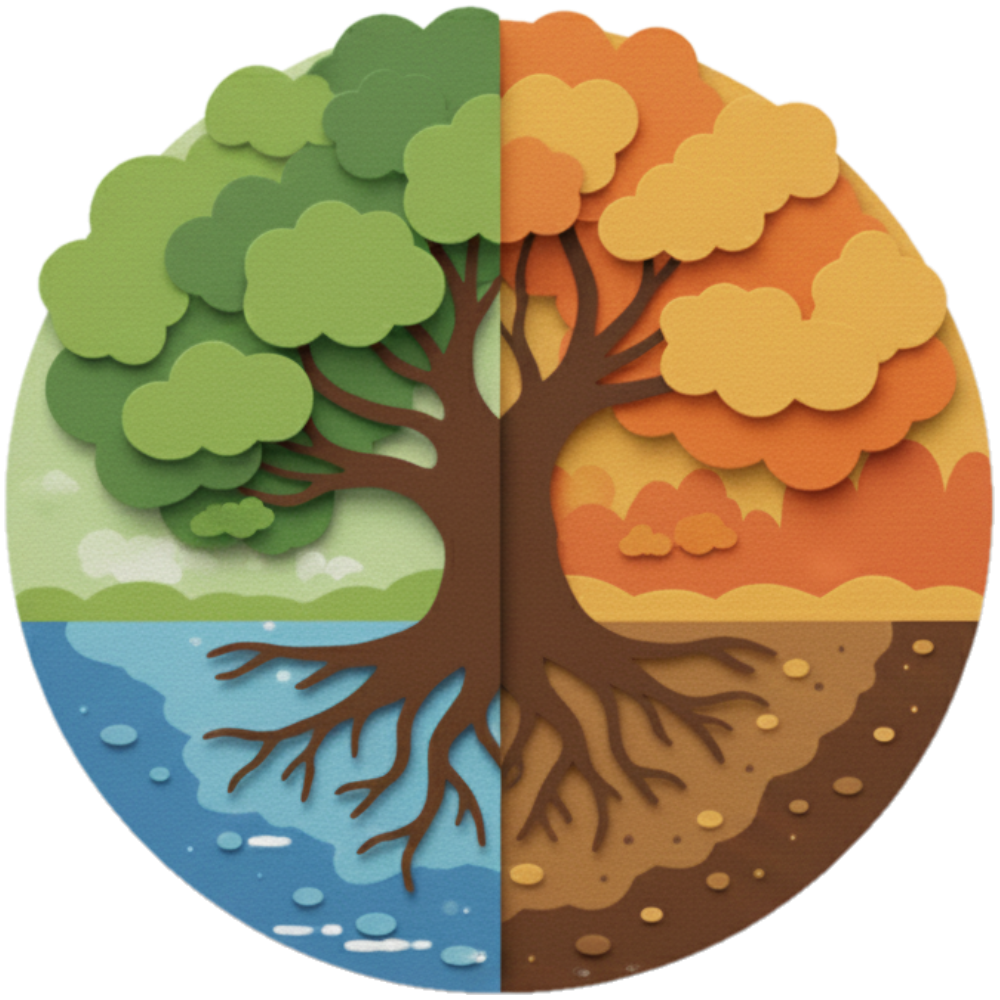} SEASON: Mitigating Temporal Hallucination in Video Large Language Models via Self-Diagnostic Contrastive Decoding}

\author{
    {Chang-Hsun Wu}$^{1, \dagger}$, {Kai-Po Chang}$^1$, {Yu-Yang Sheng}$^1$, \\
    {Hung-Kai Chung}$^1$, {Kuei-Chun Wang}$^1$, and {Yu-Chiang Frank Wang}$^{1,2,\ddagger}$\\
    \normalsize \textsuperscript{1} Graduate Institute of Communication Engineering, National Taiwan University  \quad
    \normalsize \textsuperscript{2} NVIDIA\\
    {\tt\small $^{\dagger}$r14942083@ntu.edu.tw, $^{\ddagger}$frankwang@nvidia.com}
}
\begin{document}

\newcommand{\methodname}{\mbox{\textit{SEASON}}\xspace}
\newcommand{\frank}[1]{\textcolor{red}{#1}}
\newcommand{\kp}[1]{\textcolor{blue}{#1}}

\maketitle
\begin{abstract}

Video Large Language Models (VideoLLMs) have shown remarkable progress in video understanding. However, these models still struggle to effectively perceive and exploit rich temporal information in videos when responding to user queries. Therefore, they often generate descriptions of events that are temporally inconsistent or causally implausible, causing severe hallucination issues. While most prior studies have focused on spatial hallucinations (e.g. object mismatches), temporal reasoning in video understanding remains relatively underexplored. To address this issue, we propose \textit{\textbf{Se}lf-Di\textbf{a}gnostic Contra\textbf{s}tive Dec\textbf{o}di\textbf{n}g (\textbf{\methodname})}, a training-free method that adaptively enhances temporal and spatial faithfulness for each output token. It achieves this by dynamically diagnosing each token's hallucination tendency and applying adaptive contrastive decoding against its corresponding temporal and spatial negatives. Extensive experiments demonstrate that \methodname outperforms all existing training-free hallucination mitigation approaches on three hallucination examination benchmarks, while further improves VideoLLMs across four general video understanding benchmarks. The code will be released upon acceptance.

\end{abstract}

    
\section{Introduction}
\label{sec:introduction}

\begin{figure}[t]
\centering
\includegraphics[width=0.475\textwidth]{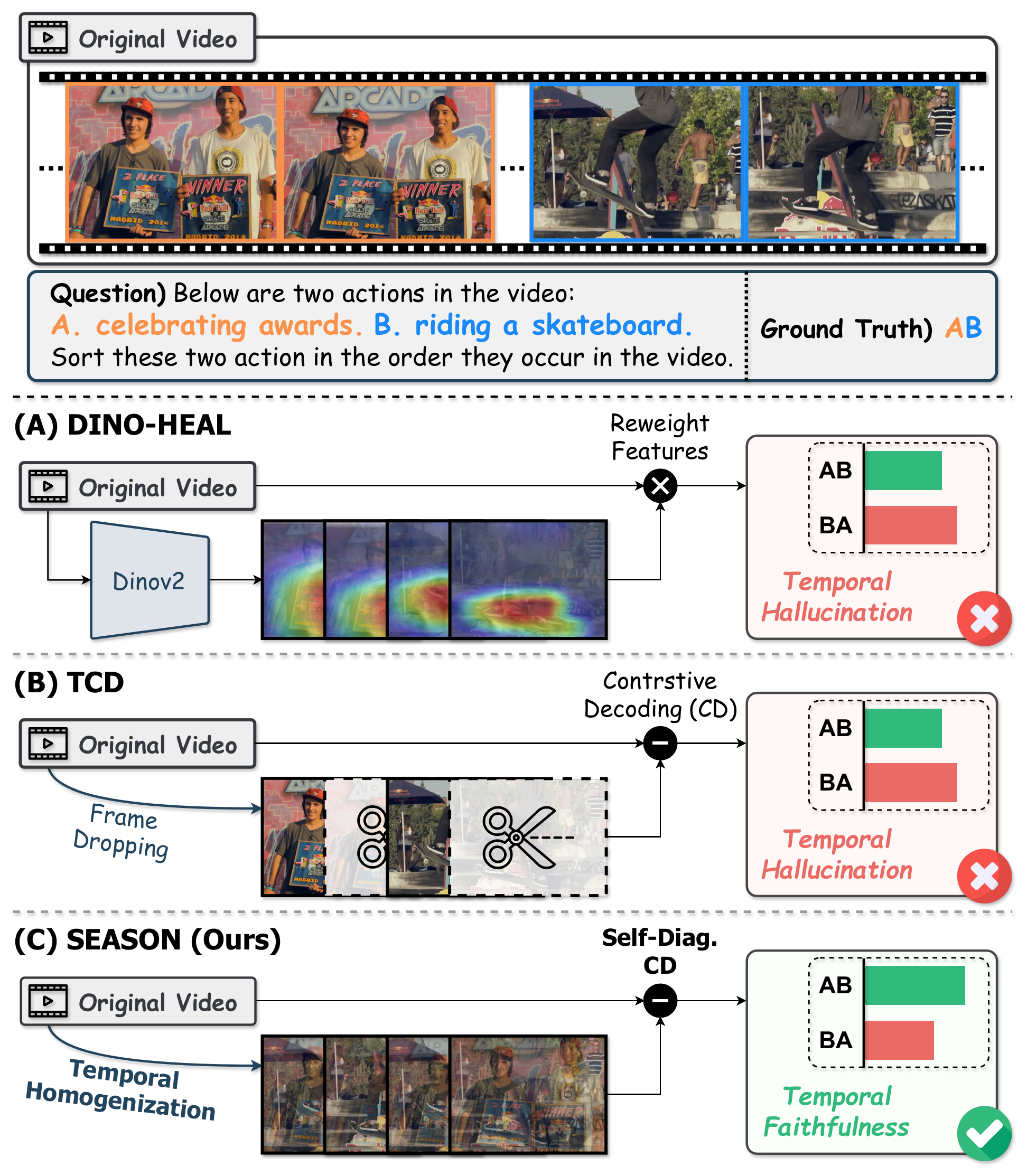}
\vspace{-7mm}
\caption{
Suppressing hallucination in video LLMs. (A) DINO-HEAL~\cite{vidhalluc} exploits spatial saliency but misses temporal order, (B) TCD~\cite{eventhallusion} contrasts frame-dropped videos but ignores causal relations, and (C) our \methodname achieves temporal faithfulness for each output token.
}
\vspace{-5mm}
\label{fig:teaser}
\end{figure}

Multimodal large language models (MLLMs)~\cite{llava, qwenvl, gpt4, gemini} have recently achieved remarkable success in bridging vision and language, enabling unified reasoning across tasks such as visual captioning~\cite{imagecap, msrvtt}, visual question answering~\cite{vqa, activitynetqa}, and video understanding~\cite{videounderstanding, videochat}. However, these models remain prone to produce the textual content that is often inconsistent with the given visual evidence, which causes severe hallucination issues~\cite{hallucination1, hallucination2}. This presents serious risks for the model deployment in critical applications demanding high-standard reliability and trustworthiness, such as healthcare robots and self-driving vehicles. As a result, alleviating hallucinations in multimodal models has become an important research focus across academia and industry.

To deal with this challenge, early works~\cite{vcd, marine, avisc, opera} have primarily focused on mitigating hallucinations in image understanding tasks, where hallucinations often appear as spatial inconsistencies, such as describing nonexistent objects or incorrect attributes~\cite{vcd}. We refer to this issue as \textit{spatial hallucination}. To handle this issue, they aim to enforce the output description exempted from spurious correlation caused by language prior, via contrasting output distributions derived from original and distorted visual inputs~\cite{avisc, opera}. Although these approaches are effective when the visual input is a \textit{static} image. However, directly extending them to video large language models is not sufficient, because videos introduce rich temporal structure. Despite reduced spatial hallucinations, the model would still misunderstand event causality and thus produce descriptions that are temporally inconsistent with visual content~\cite{videohallucer, vidhalluc}. This issue is known as \textit{temporal hallucination}, which remains a key obstacle for reliable video understanding.

To mitigate temporal alongside spatial hallucinations in video, multiple benchmarks~\cite{vidhalluc,videohallucer,eventhallusion} have been established. Building upon these, recent approaches have extended the ideas from image hallucination mitigation to video domain, generally following two research lines. Training-based methods~\cite{rrpo,arrowrl,tpo} exhibits improved temporal faithfulness via reinforcement learning~\cite{arrowrl} or preference optimization~\cite{rrpo, tpo}, but require both expensive re-training and high-quality preference data. This motivates the pursuit of training-free approaches that bypass these cost and could be easily applied to different models during inference. However, existing training-free approaches~\cite{vidhalluc, eventhallusion} such as DINO-HEAL~\cite{vidhalluc} and TCD~\cite{eventhallusion} still struggle to understand the temporal causality, as illustrated in Fig.~\ref{fig:teaser}. As a result, enabling VideoLLMs without training cost to produce descriptions that are especially temporally faithful, together with spatial fidelity, remains a challenging open problem.

In this paper, we propose \textit{\textbf{Se}lf-Di\textbf{a}gnostic Contra\textbf{s}tive Dec\textbf{o}di\textbf{n}g (\textbf{\methodname})}, a training-free method that enhances both temporal and spatial faithfulness in VideoLLMs. To specifically achieve temporal faithfulness while dynamically assessing which types of hallucination each token may suffer, we propose \textit{temporal homogenization} to construct temporally-negative video to expose the spurious temporal correlations lying within VideoLLMs, and a \textit{self-diagnostic mechanism} to identify each token's hallucination tendency based on its preceding context. More specifically, the former produces a temporal negative that is temporally incoherent yet remains spatially consistent. The latter identifies the potential hallucination type of the current token by analyzing the frame-attention divergences between the original video and these specialized negatives (temporal and spatial). By diagnosing the hallucination tendency of each token and contrasting with an appropriate negative, \methodname enables VideoLLMs to produce textual responses that are temporally faithful while maintaining spatial fidelity, without any additional training cost.

In summary, our contributions are as follows:
\begin{itemize}
  \item We present \textit{\methodname}, a self-diagnostic contrastive decoding framework that adaptively enhances temporal and spatial faithfulness for each output token in a training-free manner.
  \item \textit{\methodname} introduce temporally-homogenized videos as negatives that amplify spurious temporal correlations, yielding a temporally hallucination-focused distribution for contrastive decoding.
  \item We develop the hallucination diagnostician that estimates token-level hallucination tendency via frame-attention divergence and assigns a corresponding contrastive penalty.
\end{itemize}

\section{Related Work}
\label{sec:related_work}

\subsection{Mitigating Spatial Hallucination in Visual Large Language Models}

Visual Large Language Models (VLLMs)~\cite{llava, qwenvl, gpt4, gemini} have demonstrated remarkable capabilities in vision-language tasks. However, VLLMs often suffer from spatial hallucination, generating spatially inconsistent descriptions due to over-reliance on language priors rather than actual visual evidence~\cite{hallucination1, hallucination2}.  To mitigate these errors, research has explored two directions. Training-based approaches improve visual–language alignment through higher-quality datasets~\cite{lrvinstruction} or human feedback~\cite{rlhfv}.  While effective, these methods require large-scaled annotated datasets and costly re-training. 

On the other hand, training-free approaches~\cite{marine, vcd, avisc, opera} modify the decoding process of MLLMs without updating parameters. For example, VCD~\cite{vcd} reduces hallucination by counteracting language priors, while MARINE~\cite{marine} leverages auxiliary guidance signals to correct false information. Although efficient and widely applicable, these methods for hallucination mitigation are designed for a static image and thus cannot capture temporal dependencies within video. As directly extending to video domain, they struggle to capture temporal dynamics in video and often encounter the challenge of temporal hallucination~\cite{vidhalluc, videohallucer}.

\begin{figure*}[t]
\centering
\vspace{-6mm}
\includegraphics[width=0.975\textwidth]{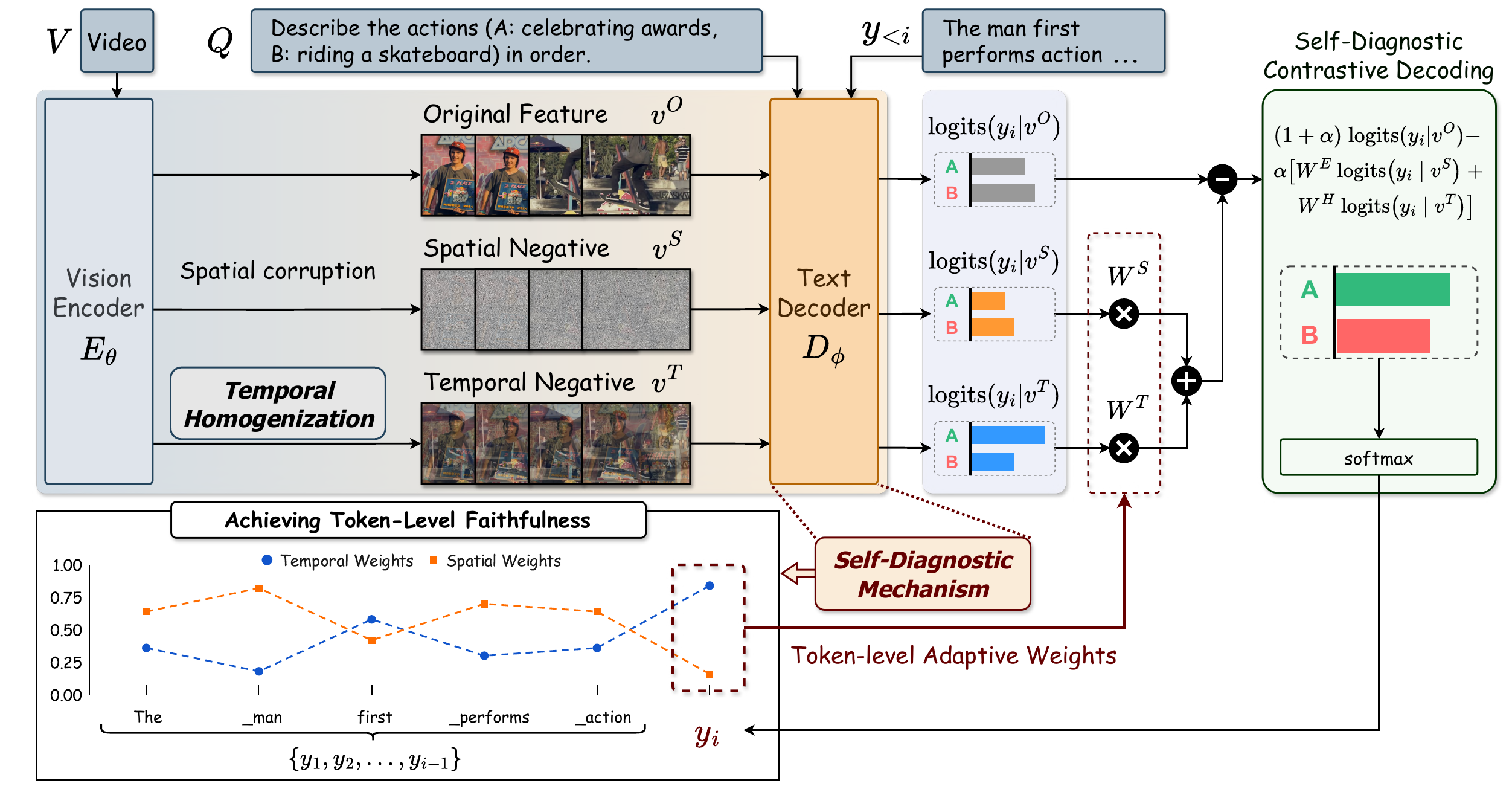}
\vspace{-2mm}
\caption{\textbf{Overview of \methodname.} Given the input video ($V$) and the question ($Q$), our proposed \methodname contrasts the original video representations ($v^O$) against our introduced spatial ($v^S$) and temporal ($v^T$) negatives to jointly achieve temporal and spatial faithfulness. Specifically, we design $v^T$ via the proposed ``Temporal Homogenization'', focusing on introducing temporal ambiguity while preserving spatial semantics. The ``Self-Diagnostic Mechanism'' computes token-level adaptive weights ($W^S, W^T$) by measuring attention divergence, dynamically steering the final decoding to penalize spatial or temporal hallucinations.
}
\vspace{-4mm}
\label{fig:mainfig}
\end{figure*}

\subsection{Mitigating Temporal Hallucination in Video Large Language Models}

The issue of hallucinations in Video Large Language Models (VideoLLMs) has emerged as a critical research area, catalyzing the development of dedicated benchmarks~\cite{videohallucer, eventhallusion, vidhalluc}. Building upon these, several works including training-based and training-free have been proposed to advance this research area. Training-based methods~\cite{arrowrl,tpo,rrpo,pamivdpo,taae,mashvlm} primarily employ reinforcement learning~\cite{arrowrl}, preference optimization~\cite{tpo,rrpo,pamivdpo}, or pre-training from scratch~\cite{mashvlm}. For example, ArrowRL~\cite{arrowrl} encourages divergent interpretations between forward and reversed videos, while RRPO~\cite{rrpo} utilizes sub-sequence-level refined rewards and a token-wise regularizer. Despite promising results, they require costly retraining, auxiliary reward models, and high-quality preference data, which limits their scalability and model-agnostic deployment.

In contrast, training-free methods bypass these costs and be able to applied to various models at inference time, offering an attractive alternative. For example, DINO-HEAL~\cite{vidhalluc} leverages saliency maps from DINOv2~\cite{dinov2} to re-weight visual features for improving object motion understanding, while TCD~\cite{eventhallusion} contrasts token predictions between original and frame-dropped videos. Nevertheless, these methods still struggle to understand complex temporal relationships, notably event causality. To address this issue, we propose \methodname to employ temporally-homogenized video as contrasted negatives, encouraging VideoLLM to understand the temporal relationship within video to achieve temporal faithfulness.

\section{Method}
\label{sec:method}

\subsection{Problem Formulation}

Given an input video $V = \{f_1, f_2, \dots, f_{|V|}\}$ consisting of $|V|$ frames and an associated textual query $Q$, a VideoLLM (parameterized by vision encoder $E_\theta$ and text decoder $D_\phi$) aims to generate a textual response $y = \{ y_1, \dots, y_N \}$ that accurately answers queries about the visual content.
A hallucination occurs when the generated response in $y$ is not presented in the video $V$. Formally, we define a \textbf{spatial hallucination} as objects or attributes described in $y$ that are visually absent within individual frames $f_i$, and a \textbf{temporal hallucination} as the descriptions in $y$ that contradict the actual temporal structure (e.g., event order and causality) presented in video $V$.

Therefore, we propose \textbf{\methodname}, a self-diagnostic contrastive decoding approach for adaptively suppressing temporal and spatial hallucinations during inference. As illustrated in Fig.~\ref{fig:mainfig}, \methodname first mitigates the temporal hallucination by providing a strong temporal negative signal to contrast (Sec.~\ref{sec:thv}). Furthermore, we introduce a self-diagnosing contrastive decoding strategy that diagnoses each token's hallucination tendency and then applies an adaptive contrastive penalty against the corresponding (temporal or spatial) negatives (Sec.~\ref{sec:selfdiag}).

\subsection{Mitigating Temporal Hallucination via Temporal Homogenization}
\label{sec:thv}
To mitigate temporal hallucination, we aim to explicitly expose and penalize a VideoLLM’s reliance on spurious temporal correlations. To achieve this, we introduce \textit{temporal homogenization}, a novel augmentation that constructs negatives which are temporally incoherent yet spatially consistent. We specifically focus on this design because simply following \citeauthor{vcd}~\cite{vcd} by adding Gaussian noise to video frames (which produces spatial negative $s^v$, Fig.~\ref{fig:mainfig}) corrupts both spatial and temporal structures simultaneously. This yields a \textit{temporally-easy} negative, as it provides suboptimal contrast by allowing the model to reject this negative based on the obvious spatial corruption rather than focusing on the intended temporal inconsistency.

In contrast, our temporal negative $v^T$ (Fig.~\ref{fig:mainfig}) maximally preserves spatial fidelity, which ensures the resulting contrastive signal reflects only temporal inconsistencies, yielding a \textit{temporally-hard} negative. This is achieved by temporally aggregating all the frame features in a layer-wise manner and respectively re-injecting them back into each frame’s representation, neutralizing temporal variation while retaining spatial semantics. See the discussion below.

\begin{figure}[t]
\centering
\vspace{-6mm}
\includegraphics[width=0.49\textwidth]{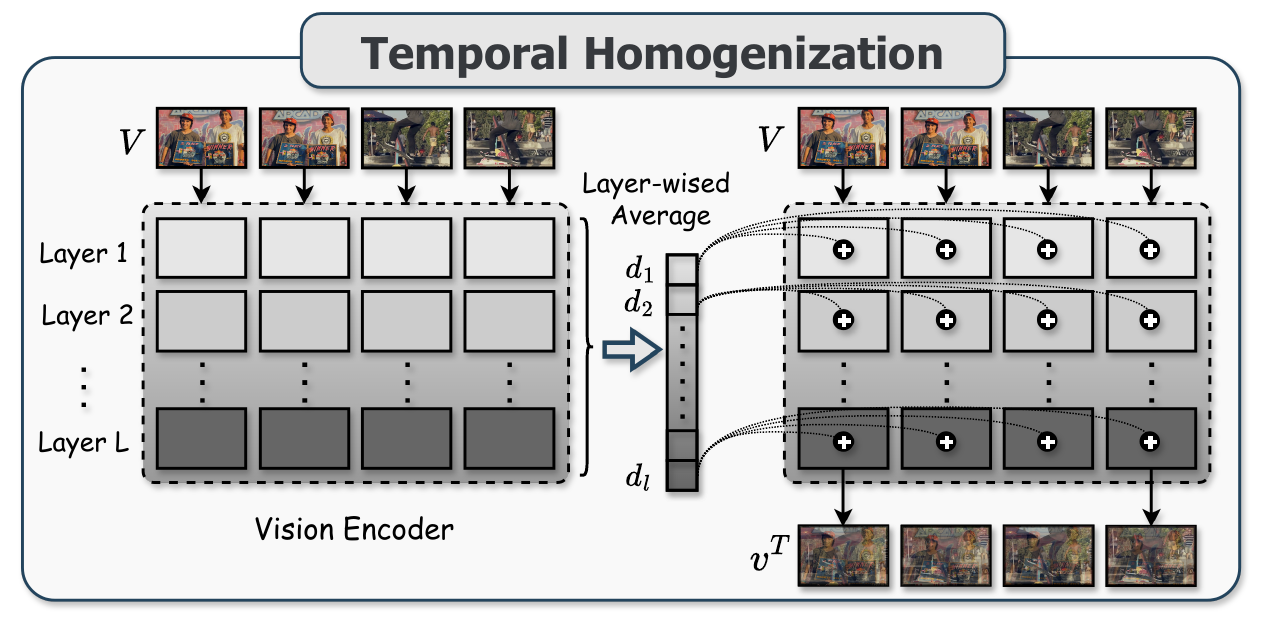}
\caption{Illustration of Temporal Homogenization. This constructs the temporal negative $v^T$ by computing a layer-wise average of frame features ($d_1,...,d_l$) and progressively re-injecting this global context back into each frame's representation within the vision encoder. The resulting representation would be temporally ambiguous while preserving per-frame structure information.}
\vspace{-2mm}
\label{fig:auxfig1}
\end{figure}


\paragraph{\mbox{Temporal Homogenization for Hallucination Exposure.}}
To expose the VideoLLM’s potential temporal hallucination, we propose the ``temporal homogenization'' to produce the temporal negatives $v^T$.  $v^T$ is an augmented video representation designed to be temporally incoherent yet spatially consistent and be constructed in Fig.~\ref{fig:auxfig1}. The core idea to achieve this is to homogenize the temporal information across the frames in a layer-wise manner while keeping per-frame spatial structure intact.

Specifically, we define the temporal negatives $v^T$ as the set of final-layer homogenized frame features from vision encoder ($E_\theta$), 
\begin{equation} 
v^T=\{h_{L,t}\}_{t=1}^{|V|}.
\end{equation} 
Each $h_{L,t}$ is obtained through a progressive and layer-wise homogenization process. At every layer $l$ and frame $t$, the homogenized feature $h_{l,t}$ is defined as a linear combination of the frame feature from the corresponding global context $d_l$ and the pre-homogenization feature $h^{\prime}_{l,t}$.  This feature $h^{\prime}_{l,t}$ is computed from the previous layer's output $h_{l-1,t}$, which has already been recurrently mixed with global context from preceding layers. 
\begin{equation} 
h_{l,t} = (1 - \beta) h^{\prime}_{l,t} + \beta d_l, \, \text{where} \; h^{\prime}_{l,t} = \text{$E_{\theta}^{(l)}$}(h_{l-1,t}). \label{eq:blend} 
\end{equation}
Here, $d_l$ denotes the mean of the frame features in $l$-th layer pre-computed from a standard forward pass on $V$ ($d_l=\frac{1}{|V|}\sum_{t=1}^{|V|}{h^{\prime}_{l,t}}$), $\beta \in [0,1]$ is a hyperparameter used to control the degree of temporal homogenization, and $h_{0,t}$ are the patch embeddings of frame $f_t$.

\paragraph{Mitigating Temporal Hallucination via Contrastive Decoding.}
With the obtained temporal negatives $v^T$, we aim to contrast the temporally hallucinated output distribution induced by $v^T$ against those from the original $v^O$. This enable us to achieve the temporal faithfulness. Thus, we impose a visual contrastive decoding to eliminate the temporal priors lying in the text decoder  ($D_\phi$). Formally, given a textual query $Q$ and the video representations $v^O$ (original) and $v^T$ (temporal negative), the contrastive distribution $p_{\textit{SEASON}^{\, \textbf{\textit{T}}}}$ is formulated as:
\begin{equation} 
\begin{aligned}
p_{\textit{SEASON}^{\, \textbf{\textit{T}}}}(y_i)  = \text{softmax}[
&(1+\alpha)\,\text{logits}(y_i|v^O,Q,y_{<i})\\ &- \alpha\,\text{logits}(y_i|v^T,Q,y_{<i})],
\label{pst}
\end{aligned}
\end{equation}
where $\alpha$ controls the contrastive strength ($\alpha=0$ reduces to regular decoding). The resulting outputs $y_i$ are generated from a distribution explicitly purified of temporal hallucination. 

Building on the formulation (Eq. ~\ref{pst}) for achieving temporal faithfulness, we next introduce our full self-diagnostic contrastive decoding strategy, a unified framework that also mitigates spatial hallucination by incorporating the spatial negatives. The effectiveness of our core temporal negatives, $v^T$, will also be experimentally verified in Table~\ref{tab:temporal-negatives}.

\begin{figure}[t]
\centering
\vspace{-6mm}
\includegraphics[width=0.49\textwidth]{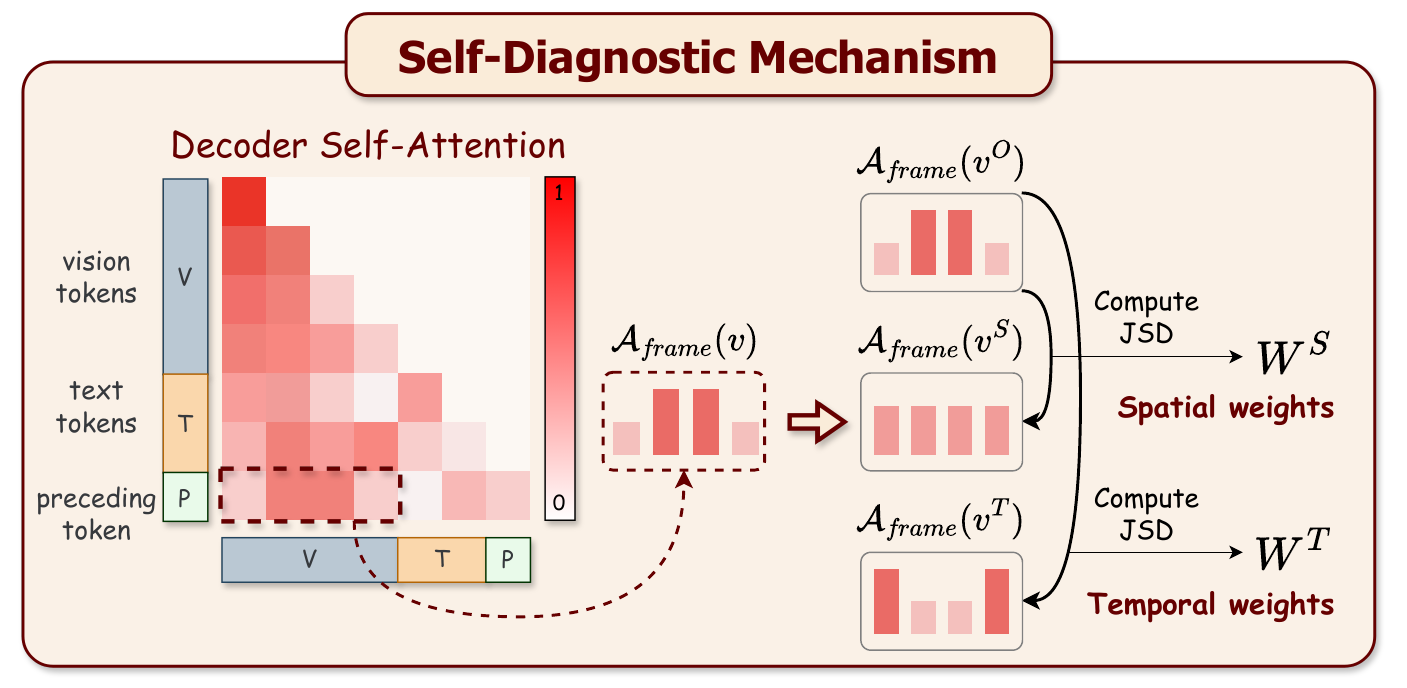}
\caption{Illustration of the Self-Diagnostic Mechanism. This process extracts the frame-level attention distribution ($\mathcal{A}_\textit{frame}$) from the preceding token. It computes JSD divergence between the attention distributions of the original video ($v^O$) and the negatives ($v^S$, $v^T$), outputting the adaptive spatial ($W^S$) and temporal ($W^T$) diagnostic weights to penalize spatial or temporal hallucination for each output token. 
}
\vspace{-2mm}
\label{fig:auxfig2}
\end{figure}

\subsection{Achieving Token-Level Faithfulness via Self-Diagnostic Contrastive Decoding}
\label{sec:selfdiag}

In Sec.~\ref{sec:thv}, the VideoLLM is able to mitigate temporal hallucination via contrastive decoding. To jointly mitigate spatial hallucination together, we propose a ``self-diagnostic mechanism'' to assess the risk of each token being temporally or spatially hallucinated. Our insight is that the generation of a token is highly dependent on its preceding context, which serves as an indicator to reflect the potential hallucination tendencies. To achieve our goal, we interpret the frame-level attention divergence between preceding text token ($y_{i-1}$) and video representation ($v\in[v^O, v^S, v^T]$) as a measure of the current token ($y_i$)’s reliance on temporal or spatial cues. For this interpretation, the required spatial negative (referenced in Fig.~\ref{fig:mainfig}) is created by adding the Gaussian noise to the original video representation $v^O$. We now explain this mechanism and how it is integrated with contrastive decoding below:

\paragraph{Detecting Hallucination Tendency via Self-Diagnostic Mechanism.}
As illustrated in Fig.~\ref{fig:auxfig2}, the self-diagnostic mechanism identifies each token’s hallucination tendency by observing how the attention pattern of its preceding token changes when temporal cues are removed (i.e., the changes between original video and temporal negatives, $v^O$ and $v^T$). The core idea is that tokens that rely on temporal consistency will exhibit strong attention shifts when the video is temporally homogenized, whereas tokens grounded in static objects remain unaffected. Consequently, the divergence between the frame-level attention distributions of the original video representations, spatial, and temporal negatives (denoted as $v^O$, $v^S$, and $v^T$, in Fig.~\ref{fig:mainfig}, respectively) serves as a self-diagnostic signal, indicating which types of hallucination (temporal or spatial) a token is prone to and how penalties should be allocated during decoding.

To quantify this behavior, we derive the frame-level attention distribution $\mathcal{A}_\textit{frame}(v)$ from the text decoder ($D_\phi$)’s multi-head attention. Specifically, $\mathcal{A}_\textit{frame}(v)$ is denoted as the normalized attention that the preceding token $y_{i-1}$ assigns to each video frame, calculated by giving the attention matrix of layer $j$ (denoted as $A_j \in \mathbb{R}^{n,n}$) obtained by summing over all heads, preceding text token $y_{i-1}$, and $v_{t,k}$ (the $k$-th visual token extracted from frame $f_t$ in the video representation $v$ ($v \in [v^O, v^S, v^T]$). This is formulated as:
\begin{equation}
\mathcal{A}_\textit{frame}(v) = \text{softmax}_t\,\Big[\! \sum_{k} \big(\!\sum_{j\in J}\! A_j\big)(y_{i-1}, v_{t,k})\!\Big].
\label{eq:attn}
\end{equation}
To diagnose the nature of current token ($y_i$)'s hallucination tendency, we compare the $\mathcal{A}_\textit{frame}(v)$ scores computed on original video representations, spatial, and temporal negatives  ($v^O$, $v^S$, and $v^T$). Specifically, we compute the divergence between these frame-level attention distributions ($\mathcal{A}_\textit{frame}(v^O)$ vs. $\mathcal{A}_\textit{frame}(v^S)$ and $\mathcal{A}_\textit{frame}(v^O)$ vs. $\mathcal{A}_\textit{frame}(v^T)$) to determine which type of hallucination the current token is prone to. Formally, let $w_S$ and $w_T$ denote the tendency to be spatially or temporally hallucinated, which are calculated via Jensen-Shannon divergences (JSD) to measure the relative degree of attention shift as follows:
\begin{equation}
\begin{aligned}
&w_S = \frac{D_S}{D_S + D_T},\quad w_T = \frac{D_T}{D_S + D_T}, \\
&D_S = \text{JSD}(\mathcal{A}_\textit{frame}(v^O),\mathcal{A}_\textit{frame}(v^S)), \\ 
&D_T = \text{JSD}(\mathcal{A}_\textit{frame}(v^O),\mathcal{A}_\textit{frame}(v^T)).
\end{aligned}
\label{eq:diagnosis}
\end{equation}
Here, $D_S$ and $D_T$ measure the relative degree of attention shift. A larger $D_T$ signifies a temporal hallucination tendency (as the token relies heavily on temporal cues), while a larger $D_S$ signifies a spatial hallucination tendency.

By this method, the derived adaptive weights $(w_S, w_T)$ thus serve as the core diagnostic signals, quantifying the token's specific hallucination tendency and establishing the basis for the subsequent contrastive decoding stage.

\paragraph{Self-Diagnostic Contrastive Decoding.} 
To mitigate temporal alongside spatial hallucinations of VideoLLM without any further training cost, we integrate the self-diagnostic mechanism into contrastive decoding. The adaptive diagnostic weights $(w_S, w_T)$ derived above are thus incorporated to dynamically balance the contrastive penalties at each generation step.

Given the current decoding step $i$, we first obtain three logit distributions by feeding the textual context $(y_{<i}, q)$ to the text decoder $D_\phi$ conditioned on the original video representation $v^O$, spatial negative $v^S$, and temporal negative $v^T$. For brevity, we denote these logits distributions as $\text{logits}(y_i|v^O)$, $\text{logits}(y_i|v^S)$, and $\text{logits}(y_i|v^T)$, respectively. Therefore, the final self-diagnostic contrastive decoding distribution $p_{\textit{SEASON}}$ that adaptively suppresses both spatial and temporal hallucinations is formulated, by combining these logits using the diagnostic weights:
\begin{equation}\begin{aligned}
p_{\textit{SEASON}}(y_i)
= \text{softmax}\Big[(1 + \alpha)\,\text{logits}(y_i|v^O)\\
- \alpha \, [w_S\,\text{logits}(y_i|v^S) + w_T \,\text{logits}(y_i|v^T)\Big].
\label{eq:sdd}
\end{aligned}
\end{equation}
This formulation acts as logit-space contrastive decoding, where the adaptive weights $(w_S, w_T)$ determine the penalty for potential spatial or temporal hallucination tendency for each token. A larger $w_T$ suppresses potential temporal hallucination, while a larger $w_S$ penalizes possible spatial hallucination. 

By dynamically steering the decoding direction in this manner, the model achieves per-token self-assessment and adaptively corrects potential hallucinations, ensuring both temporal and spatial faithfulness throughout the generation process.

\section{Experiments}
\label{sec:experiments}

\newcommand{\cmark}{\textbf{\textcolor{green!60!black}{\ding{51}}}}
\newcommand{\xmark}{\textbf{\textcolor{red!70!black}{\ding{55}}}}

\begin{table*}[t]
\centering
\vspace{-5mm}
\caption{Evaluation of multiple hallucination examination benchmarks with different VideoLLMs as backbones. \textbf{Bold} marks the best per group; highlights indicate the top two benchmark results.}
\vspace{-3mm}
\label{tab:halbanchmark}
\resizebox{\textwidth}{!}{%
\begin{tabular}{lccccccccccccc}
\toprule
\multirow{2}{*}{\textbf{Models}} & \multirow{2}{*}{\textbf{\shortstack[c]{Training-\\free}}}
& \multicolumn{5}{c}{\textbf{VidHalluc}}
& \multicolumn{6}{c}{\textbf{VideoHallucer}}
& \multicolumn{1}{c}{\textbf{EventHallusion}} \\
\cmidrule(lr){3-7} \cmidrule(lr){8-13} \cmidrule(lr){14-14}
& & BQA & MCQ & STH & TSH & AVG
& ORH & TPH & SDH & EFH & ENFH & AVG
& AVG \\
\midrule

LLaVA-OV-7B~\cite{llavaov} & \textbf{-} & \textbf{74.36} & 90.27 & \textbf{63.65} & 53.00 & 70.32 & 56.50 & 52.50 & \textbf{56.50} & 15.00 & 51.50 & 46.40 & 60.15 \\
+TCD~\cite{eventhallusion} & \cmark & 72.44 & 90.06 & 58.57 & 64.33 & \cellcolor{blue!10}71.35 & 59.50 & 53.50 & 56.00 & 17.50 & \textbf{54.00} & \cellcolor{blue!10}48.10 & \cellcolor{blue!10}68.46 \\
+DINO-HEAL~\cite{vidhalluc} & \cmark & 74.29 & 90.36 & 63.18 & 53.00 & 70.21 & 57.00 & 53.50 & \textbf{56.50} & 15.50 & 52.00 & 46.90 & 60.15 \\
\textbf{+\methodname (Ours)} & \cmark & 73.15 & \textbf{90.51} & 60.29 & \textbf{77.50} & \cellcolor{blue!20}\textbf{75.36} & \textbf{63.00} & \textbf{55.50} & \textbf{56.50} & \textbf{19.50} & 48.00 & \cellcolor{blue!20}\textbf{48.50} & \cellcolor{blue!20}\textbf{69.19} \\

\midrule

QWEN2.5-VL-7B~\cite{qwen25vl} & \textbf{-} & 75.79 & 84.05 & 74.91 & 59.00 & 73.44 & 62.00 & 46.50 & 70.50 & 31.50 & 55.00 & 53.10 & 63.33 \\
+TCD~\cite{eventhallusion} & \cmark & 74.60 & 85.57 & 74.37 & 64.67 & \cellcolor{blue!10}74.80 & 61.00 & 46.50 & 70.50 & 31.00 & 55.00 & 52.80 & 64.79 \\
+DINO-HEAL~\cite{vidhalluc} & \cmark & 75.86 & 84.21 & \textbf{75.86} & 58.67 & 73.65 & 61.00 & 46.50 & 71.50 & 32.00 & 55.50 & 53.30 & 63.57 \\
+ArrowRL~\cite{arrowrl} & \xmark & 76.14 & \textbf{87.84} & 70.92 & 57.83 & 73.18 & 60.50 & \textbf{55.00} & 68.00 & \textbf{33.50} & 55.50 & \cellcolor{blue!10}54.50 & \cellcolor{blue!20}\textbf{68.95} \\
\textbf{+\methodname (Ours)} & \cmark & \textbf{78.08} & 87.32 & 71.92 & \textbf{77.67} & \cellcolor{blue!20}\textbf{78.75} & \textbf{63.50} & 49.50 & \textbf{73.50} & 31.50 & \textbf{56.00} & \cellcolor{blue!20}\textbf{54.80} & \cellcolor{blue!10}66.01 \\

\midrule

LLaVA-Video-7B~\cite{llavavideo} & \textbf{-} & 75.02 & 90.76 & 51.23 & 38.00 & 63.75 & 60.00 & 61.50 & 66.50 & 16.50 & \textbf{52.50} & \cellcolor{blue!20}\textbf{51.40} & 63.57 \\
+TCD~\cite{eventhallusion} & \cmark & 73.50 & 90.41 & 49.64 & 46.00 & \cellcolor{blue!10}64.89 & 58.00 & 61.00 & 65.50 & 14.50 & 51.50 & 50.10 & 64.30 \\
+DINO-HEAL~\cite{vidhalluc} & \cmark & 75.27 & 90.81 & \textbf{51.51} & 37.67 & 63.81 & 59.50 & 61.00 & 66.50 & 17.00 & \textbf{52.50} & 51.30 & 64.30 \\
+TPO~\cite{tpo} & \xmark & 74.85 & 90.69 & 49.62 & 42.50 & 64.42 & 60.00 & 59.50 & \textbf{68.50} & 16.00 & \textbf{52.50} & 51.30 & 63.33 \\
+RRPO~\cite{rrpo} & \xmark & \textbf{76.80} & \textbf{91.23} & 49.83 & 37.67 & 63.88 & 59.00 & 58.00 & 67.00 & \textbf{21.00} & 51.50 & 51.30 & \cellcolor{blue!20}\textbf{67.97} \\
\textbf{+\methodname (Ours)} & \cmark & 74.71 & 90.95 & 49.86 & \textbf{50.33} & \cellcolor{blue!20}\textbf{66.46} & \textbf{60.50} & \textbf{62.00} & 68.00 & 18.00 & 48.50 & \cellcolor{blue!20}\textbf{51.40} & \cellcolor{blue!10}66.99 \\

\bottomrule
\end{tabular}%
\vspace{-2mm}
}
\end{table*}

\begin{table*}[t]
\centering
\caption{Performance comparisons on benchmarks for hallucination examination, temporal, and conventional video understanding. Different VideoLLMs are applied as backbones. \textbf{Bold} marks the best per group; highlights indicate the best benchmark results.
}
\vspace{-3mm}
\label{tab:allbenchmarks}
\resizebox{\textwidth}{!}{%
\begin{tabular}{lccccccccccc}
\toprule

\multirow{2}{*}{\textbf{Models}} & \multirow{2}{*}{\textbf{\shortstack[c]{Training-\\free}}}
& \multicolumn{4}{c}{\textbf{Hallucination Examination}}
& \multicolumn{3}{c}{\textbf{Temporal Understanding}}
& \multicolumn{3}{c}{\textbf{Conventional Video Understanding}} \\
\cmidrule(lr){3-6} \cmidrule(lr){7-9} \cmidrule(lr){10-12}

& & \textbf{VidHalluc} & \textbf{VideoHallucer} & \textbf{EventHallusion} & \textbf{AVG} & \textbf{TempCompass} & \textbf{TVBench} & \textbf{AVG} & \textbf{VideoMMe} & \textbf{MVBench} & \textbf{AVG}

\\

\midrule
\multicolumn{9}{l}{\textit{Proprietary models}} \\
\rowcolor{gray!8}
GPT-4o~\cite{gpt4} & \textbf{-} & 81.2 & 53.3 & 91.9 & 75.5 & 73.8 & 39.9 & 56.9 & 71.9 & 49.1 & 60.5 \\
\rowcolor{gray!8}
Gemini 1.5 Pro~\cite{gemini} & \textbf{-} & 72.8 & 37.8 & 80.4 & 63.8 & 67.1 & 47.6 & 57.4 & 75.0 & 60.5 & 67.8 \\

\midrule

LLaVA-OV-7B~\cite{llavaov} & \textbf{-} & 70.3 & 46.4 & 60.2 & 59.0 & 68.3 & 42.8 & 55.6 & 50.6 & 54.4 & 52.5 \\
+TCD~\cite{eventhallusion} & \cmark & 71.4 & 48.1 & 68.5 & 62.6 & 68.5 & 42.9 & 55.7 & 50.4 & 54.3 & 52.4 \\
+DINO-HEAL~\cite{vidhalluc} & \cmark & 70.2 & 46.9 & 60.2 & 59.1 & 68.3 & 42.9 & 55.6 & 50.6 & 54.4 & 52.5 \\
\textbf{+\methodname (Ours)} & \cmark & \textbf{75.4} & \textbf{48.5} & \textbf{69.2} & \cellcolor{blue!20}\textbf{64.3} & \textbf{69.3} & \textbf{43.0} & \cellcolor{blue!20}\textbf{56.1} & \textbf{50.7} & \textbf{54.8} & \cellcolor{blue!20}\textbf{52.7} \\

\midrule

QWEN2.5-VL-7B~\cite{qwen25vl} & \textbf{-} & 73.4 & 53.1 & 63.3 & 63.3 & 72.9 & 46.7 & 59.8 & 49.9 & 61.8 & 55.8 \\
+TCD~\cite{eventhallusion} & \cmark & 74.8 & 52.8 & 64.8 & 64.1 & 73.4 & 46.1 & 59.8 & 49.4 & 60.1 & 54.7 \\
+DINO-HEAL~\cite{vidhalluc} & \cmark & 73.7 & 53.3 & 63.6 & 63.5 & 73.2 & 46.9 & 60.1 & 49.9 & 61.9 & 55.9 \\
+ArrowRL~\cite{arrowrl} & \xmark & 73.2 & 54.5 & \textbf{69.0} & 65.5 & 72.6 & \textbf{49.4} & \cellcolor{blue!20}\textbf{61.0} & 49.6 & 59.8 & 54.7 \\
\textbf{+\methodname (Ours)} & \cmark & \textbf{78.7} & \textbf{54.8} & 66.0 & \cellcolor{blue!20}\textbf{66.5} & \textbf{73.7} & 47.7 & 60.7 & \textbf{50.4} & \textbf{62.3} & \cellcolor{blue!20}\textbf{56.4} \\

\midrule

LLaVA-Video-7B~\cite{llavavideo} & \textbf{-} & 63.8 & \textbf{51.4} & 63.6 & 59.6 & 69.7 & 45.2 & 57.5 & 53.0 & \textbf{58.4} & 55.7 \\
+TCD~\cite{eventhallusion} & \cmark & 64.9 & 50.1 & 64.3 & 59.8 & 70.0 & 45.2 & 57.6 & 53.0 & 57.5 & 55.3 \\
+DINO-HEAL~\cite{vidhalluc} & \cmark & 63.8 & 51.3 & 64.3 & 59.8 & 69.9 & 45.5 & 57.7 & 53.2 & 58.3 & 55.8 \\
+TPO~\cite{tpo} & \xmark  & 64.4 & 51.3 & 63.3 & 59.7 & 69.5 & 45.1 & 57.3 & 53.0 & 58.2 & 55.6 \\
+RRPO~\cite{rrpo} & \xmark & 63.9 & 51.3 & \textbf{68.0} & 61.1 & 70.3 & 45.2 & 57.8 & \textbf{53.8} & 58.1 & \cellcolor{blue!20}\textbf{56.0} \\
\textbf{+\methodname (Ours)} & \cmark & \textbf{66.5} & \textbf{51.4} & 67.0 & \cellcolor{blue!20}\textbf{61.6} & \textbf{71.1} & \textbf{46.4} & \cellcolor{blue!20}\textbf{58.8} & 53.4 & 57.9 & 55.6 \\
\bottomrule

\end{tabular}%
\vspace{-8mm}
}
\end{table*}
\begin{figure*}[t]
    \centering
    \vspace{-2mm}
    \includegraphics[width=\linewidth]{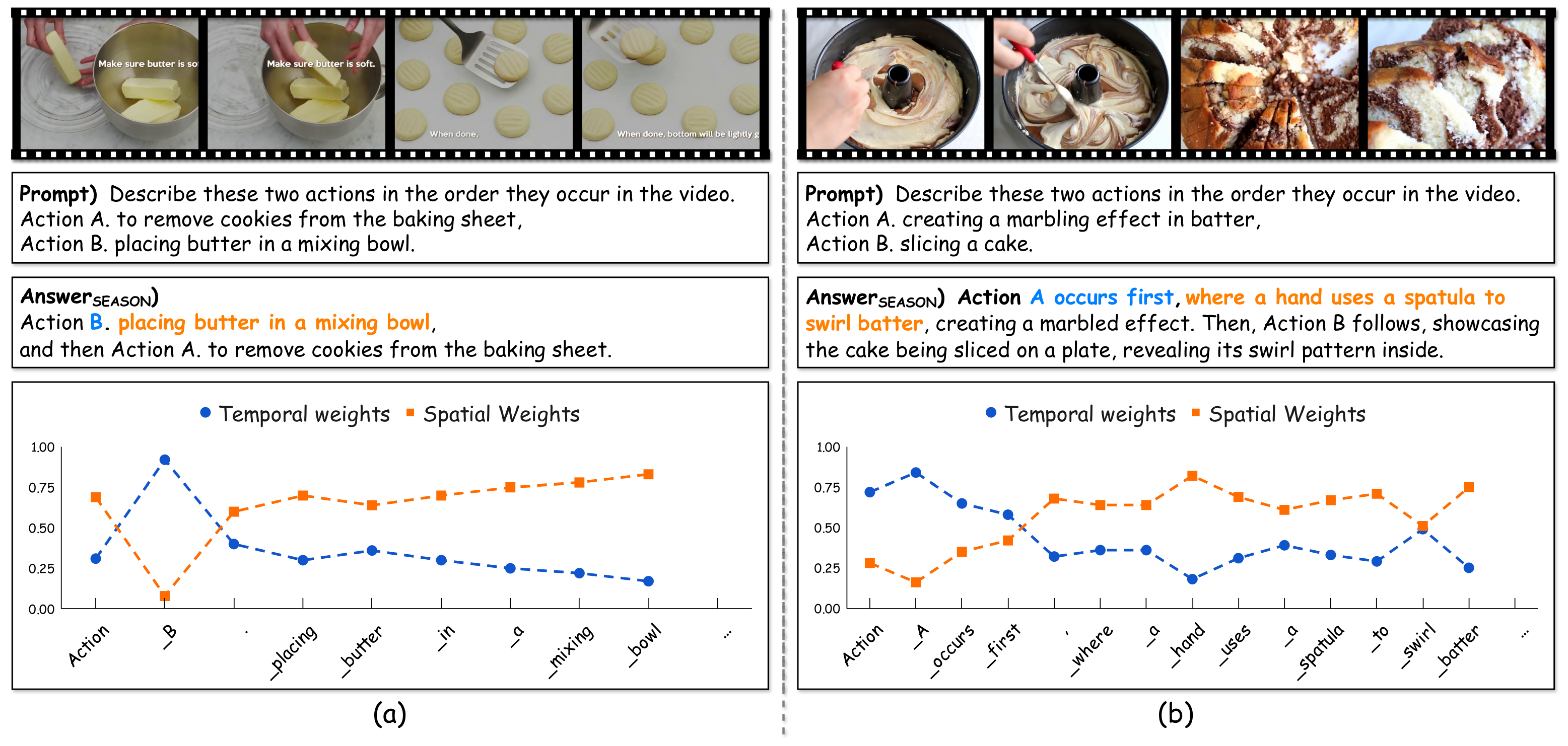}
    \vspace{-6mm}
    \caption{\textbf{Qualitative visualization of \methodname's self-diagnostic mechanism.} Qualitative visualization of \methodname's self-diagnostic weights (\textcolor{blue}{$W^T$} and \textcolor{orange}{$W^S$}). In the generated text (the x-axis in the line plot), \textcolor{blue}{blue} tokens are identified as relying on visual temporal cues; \methodname thus contrasts them against the temporal negative ($v^T$) to ensure token-level temporal faithfulness.  For instance, tokens critical for temporal ordering like ``B'' (in (a)), as well as ``A'' and ``first'' ((in (b))) clearly receive high temporal weights (\textcolor{blue}{$W^T$}) to ensure the sequence is correct. On the other hand, \textcolor{orange}{orange} tokens rely on visual spatial cues and are contrasted against the spatial negative ($v^S$). This is evident as tokens describing objects and interactions, such as ``placing butter...mixing bowl'' in (a) and ``hand...swirl batter'' in (b), are assigned high spatial weights (\textcolor{orange}{$W^S$}).
    Both (a) and (b) are samples from Vidhalluc~\cite{vidhalluc}.
    }
    \label{fig:qualitative}
  \vspace{-2mm}
\end{figure*}


\subsection{Experimental Setup}
\label{sec:setup}

\paragraph{Benchmarks and Metrics.}
To directly assess the reduction of spatial and temporal hallucinations, we evaluate \methodname on three dedicated video hallucination examination benchmarks: VidHalluc~\cite{vidhalluc}, VideoHallucer~\cite{videohallucer}, and EventHallusion~\cite{eventhallusion}.
To verify that our method preserves general video understanding, we further evaluate performance on two temporal understanding benchmarks: TempCompass~\cite{tempcompass} and TVBench~\cite{tvbench}, along with two conventional video understanding benchmarks: VideoMME~\cite{videomme} and MVBench~\cite{mvbench}.
Most subtasks are evaluated using QA accuracy following their official protocols. Key exceptions include the STH subtask in VidHalluc, which combines a classification score and a description score.
For EventHallusion~\cite{eventhallusion} and TempCompass~\cite{tempcompass}, we follow~\cite{taae, arrowrl} and reproduce only the deterministic subtasks to avoid reliance on third-party LLM-based evaluators and reduce evaluation ambiguity.
Please refer to the Appendix for more experiments and related details.

\paragraph{Base Models and Baselines.}
We apply our training-free framework to three open-source VideoLLMs (LLaVA-OV-7B~\cite{llavaov}, Qwen2.5-VL-7B~\cite{qwen25vl}, and LLaVA-Video-7B~\cite{llavavideo}) to demonstrate broad applicability.
For comparison, we include two training-free hallucination mitigation methods designed for VideoLLMs, TCD~\cite{eventhallusion} and DINO-HEAL~\cite{vidhalluc}.
In addition, as a reference for state-of-the-art performance, we report results from three training-based methods (ArrowRL~\cite{qwen25vl}, TPO~\cite{tpo}, and RRPO~\cite{rrpo}) that aim to enhance temporal reasoning ability of VideoLLMs.
To isolate the contribution of our novel components, we also perform a comparison of each negative in Tab.~\ref{tab:easy-hard}.

\paragraph{Implementation Details.}
\methodname is applied purely during inference without any retraining or fine-tuning.
We utilize 8 frames for inference, and all experiments are conducted on the same model backbones with identical settings across baselines to ensure a fair comparison.
For the self-diagnostic mechanism, we select attention layers $J = [20, 21, 22, 23]$ (Eq.~\ref{eq:attn}) based on empirical analysis.
Hyperparameters (contrastive strength $\alpha$ and homogenization degree $\beta$) are tuned via grid search to systematically explore critical settings. Additional experimental details are provided in the Appendix.


\subsection{Quantitative Evaluation}
\label{sec:quantitative}

\paragraph{Hallucination Benchmark Evaluation.}
As shown in Tab.~\ref{tab:halbanchmark}, \methodname consistently achieves the best training-free performance across all three hallucination examination benchmarks for each backbone, while remaining competitive with or even surpassing training-based methods.
For example, on QWEN2.5-VL-7B~\cite{qwen25vl}, \methodname improves the overall VidHalluc~\cite{vidhalluc} score by +5.3\% over the base model and +5.6\% over the training-based baseline (ArrowRL~\cite{arrowrl}).
The improvement is particularly pronounced in mitigating temporal hallucinations, which our method is designed to target (Sec.~\ref{sec:thv}). 
For instance, on the TSH subtask of VidHalluc~\cite{vidhalluc}, \methodname boosts performance by up to +24.5\%/+18.7\%/+12.3\% over each backbones.

\vspace{-2mm}
\paragraph{General Video Understanding Evaluation.}
Crucially, as shown in Tab.~\ref{tab:allbenchmarks}, these gains in faithfulness do not come at the cost of general comprehension. 
\methodname improves performance on the two temporal understanding benchmarks while maintaining performance on the two conventional video understanding benchmarks (e.g., +1.4\%/+1.2\% on TempCompass~\cite{tempcompass} and TVBench~\cite{tvbench} for LLaVA-Video-7B~\cite{llavavideo}). 
This result indicates that our self-diagnostic mechanism (Sec.~\ref{sec:selfdiag}) effectively penalizes suspected hallucinations without over-suppressing correct tokens.

\begin{table}[t]
  \vspace{-2mm}
  \centering
  \caption{Ablation of \methodname's temporal negative ($v^T$) design. We compare our \textbf{homogenized} strategy against alternatives (Average, Shuffled, Reverse) on temporal hallucination examination and temporal understanding benchmarks.}
  \vspace{-3mm}
  \label{tab:temporal-negatives}
  \resizebox{0.475\textwidth}{!}{%
    \begin{tabular}{lccccc}
      \toprule
      \multirow{2}{*}{\textbf{Models}} 
      & \multicolumn{2}{c}{\textbf{Temporal Hallu.}} 
      & \multirow{2}{*}{\textbf{\shortstack[l]{Temp\\Compass}}} 
      & \multirow{2}{*}{\textbf{\shortstack[l]{TV\\Bench}}} 
      & \multirow{2}{*}{\textbf{AVG}} 
      \\
      \cmidrule(lr){2-3} 
       
      & TSH & TPH 
      \\
      
      \midrule
      LLaVA-OV-7B~\cite{llavaov} & 53.0 & 52.5 & 68.3 & 42.8 & 54.2\\
      +Average & \underline{72.7} & \underline{55.5} & 69.3 & 42.2 & \underline{59.9}\\
      +Shuffled  & 52.3 & 54.5 & \textbf{69.7} & \underline{43.6} & 55.0\\
      +Reverse & 58.3 & \textbf{58.5} & 69.1 & \textbf{44.3} & 57.6\\
      \textbf{+Homogenized (ours)} & \textbf{77.7} & \underline{55.5} & \underline{69.5} & 42.9 & \cellcolor{blue!20}\textbf{61.4}\\

      \midrule
      QWEN2.5-VL-7B~\cite{qwen25vl} & 59.0 & 46.5 & \underline{72.9} & 46.7 & 56.3\\
      +Average & \underline{74.7} & 44.0 & 71.6 & 43.9 & 58.6\\
      +Shuffled  & 68.5 & \underline{50.0} & \underline{72.9} & \textbf{47.5} & \underline{59.7}\\
      +Reverse & 67.3 & \textbf{51.5} & 71.6 & 44.9 & 58.8\\
      \textbf{+Homogenized (ours)} & \textbf{79.3} & 47.0 & \textbf{74.1} & \textbf{47.5} & \cellcolor{blue!20}\textbf{62.0}\\
      
      \bottomrule
    \vspace{-2mm}
    \end{tabular}%
  }
\vspace{-6mm}
\end{table}

\subsection{Qualitative Evaluation}
\label{sec:qualitative}

To illustrate how \methodname 
mitigates hallucinations, Fig.~\ref{fig:qualitative} provides two qualitative examples. 
We visualize the token-wise adaptive weights computed by our self-diagnostic mechanism (Eq.~\ref{eq:diagnosis}). When the model generates temporal-related words (e.g., "A occurs first" in (b)), the temporal weight $w_T$ (\textcolor{blue}{blue}) is high, indicating a strong contrast against the temporal negative $v^T$. In contrast, when generating tokens for objects or static attributes (e.g., "butter" and "mixing bowl" in (a)), the spatial weight $w_S$ (\textcolor{orange}{orange}) is high, activating a contrast against the spatial negative $v^S$.

\begin{table}[t]
  \centering
  \vspace{-3mm}
  \caption{Ablation of \methodname's key components ($v^S$, $v^T$) across various hallucination examination benchmarks.
}
  \vspace{-3mm}
  \label{tab:single-ablation}
  \resizebox{0.475\textwidth}{!}{%
    \begin{tabular}{lcccc}
      \toprule
      \textbf{Models} & \textbf{VidHalluc} & \textbf{VideoHallucer} & \textbf{EventHallusion} & \textbf{AVG} \\
      
      \midrule
      LLaVA-OV-7B~\cite{llavaov} & 70.3 & 46.4 & 60.2 & 59.0\\
      +Spatial Neg. ($v^S$) & 75.1 & 46.7 & 69.2 & 63.7\\
      +Temporal Neg. ($v^T$) & 74.2 & 47.5 & \textbf{69.9} & 63.9\\
      \textbf{+\methodname} & \textbf{75.4} & \textbf{48.5} & 69.2 & \cellcolor{blue!20}\textbf{64.4}\\

      \midrule
      QWEN2.5-VL-7B~\cite{qwen25vl} & 73.4 & 53.1 & 63.3 & 63.3\\
      +Spatial Neg. ($v^S$) & 77.4 & 53.4 & 63.3 & 64.7\\
      +Temporal Neg. ($v^T$) & 77.2 & 51.9 & \textbf{68.0} & 66.2\\
      \textbf{+\methodname} & \textbf{78.7} & \textbf{54.8} & 66.0 & \cellcolor{blue!20}\textbf{66.5}\\
      
      \bottomrule
    \end{tabular}%
  }
\end{table}  

\begin{table}[t]
  \centering
  \caption{Ablation study of \methodname's key components. We evaluate the impact of the spatial negative ($v^S$) and temporal negative ($v^T$) on temporal and overall hallucination examination tasks.}
  \vspace{-3mm}
  \label{tab:easy-hard}
  \resizebox{0.475\textwidth}{!}{%
    \begin{tabular}{lcccc}
      \toprule
      \multirow{2}{*}{\textbf{Models}} 
      & \multicolumn{2}{c}{\textbf{VidHalluc}} 
      & \multicolumn{2}{c}{\textbf{VideoHallucer}} \\
      \cmidrule(lr){2-3} \cmidrule(lr){4-5}
       
      & \shortstack[c]{Temporal}
      & \shortstack[c]{Overall}
      & \shortstack[c]{Temporal}
      & \shortstack[c]{Overall}
      \\
      
      \midrule
      LLaVA-OV-7B~\cite{llavaov} & 53.0 & 70.3 & 52.5  & 46.4 \\
      +Spatial Neg. ($v^S$) & 75.7 & 75.1 & 52.0  & 46.7 \\
      +Temporal Neg. ($v^T$) & \textbf{77.7} & 74.2 & \textbf{55.5}  & 47.5 \\
      \textbf{+\methodname} & 77.5 & \textbf{75.4} & \textbf{55.5} & \textbf{48.5} \\

      \midrule
      QWEN2.5-VL-7B~\cite{qwen25vl} & 59.0 & 73.4 & 46.5  & 53.1 \\
      +Spatial Neg. ($v^S$) & 77.0 & 77.4 & 42.0  & 53.4 \\
      +Temporal Neg. ($v^T$) & \textbf{79.3} & 77.7 & 47.0  & 52.9 \\
      \textbf{+\methodname} & 77.7 & \textbf{78.7} & \textbf{49.5}  & \textbf{54.8} \\
      
      \bottomrule
    \end{tabular}%
    \vspace{-2mm}
  }
\vspace{-3mm}
\end{table}

\subsection{Ablation Study}
\label{sec:ablation}

\paragraph{Analysis on Temporal Negative.} 
To validate the design of our \textit{temporally-hard} negative (Sec.~\ref{sec:thv}), we conduct an ablation study in Tab.~\ref{tab:temporal-negatives}. We compare against three simpler \textit{temporally-easy} negatives: Average, Shuffled, and Reverse.
The results clearly demonstrate the superiority of our approach. Our Homogenized method achieves the highest overall Average score for both backbones (e.g., +7.2\% on LLaVA-OV-7B~\cite{llavaov} and +5.7\% on QWEN2.5-VL-7B~\cite{qwen25vl}).

\vspace{-2mm}
\paragraph{Ablation on Key Components.}

We analyze the contribution of our spatial ($v^S$) and temporal ($v^T$) negatives in Tab.~\ref{tab:single-ablation} and~\ref{tab:easy-hard}. From Tab.~\ref{tab:single-ablation}, while applying each of the two introduced negatives individually improves performance, their combination in \methodname achieves the highest AVG score (e.g., +5.4\% on LLaVA-OV-7B~\cite{llavaov}). In Tab.~\ref{tab:easy-hard}, the baseline of solely applying $v^T$ achieves the best temporal-only results on VidHalluc~\cite{vidhalluc}. On the other hand, \methodname surpasses this baseline on VideoHallucer~\cite{videohallucer}. We attribute this discrepancy to VideoHallucer's Yes/No VQA data format, which is susceptible to compliance bias (preference for "Yes"), as also empirically verified by \citeauthor{videohallucer}~\cite{videohallucer}. Despite this, \methodname consistently achieves the best overall (spatial and temporal) mitigation across both benchmarks, and ranks among the top two for temporal-only mitigation.

\begin{figure}[t]
\vspace{-6mm}
\centering
\includegraphics[width=0.49\textwidth]{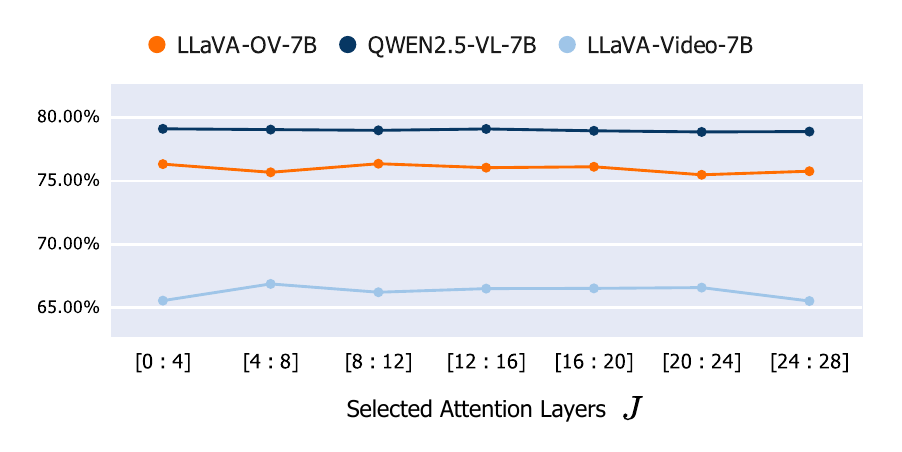}
\vspace{-8mm}
\caption{
Ablation study on the selected attention layers ($J$) for \methodname, evaluated on the VidHalluc benchmark~\cite{vidhalluc}. The performance remains robust and stable, demonstrating insensitivity to the specific layers chosen for aggregation.
}
\label{fig:attnablation}
\end{figure}

\vspace{-2mm}
\paragraph{Analysis of Self-diagnostic Layers.}
We analyze the sensitivity of our self-diagnostic mechanism (Sec~\ref{sec:selfdiag}) to the selected attention layers $J$ (Eq.~\ref{eq:attn}). In Fig.~\ref{fig:attnablation}, we plot the performance in VidHalluc~\cite{vidhalluc} over three backbone models when using different attention layers in our self-diagnostic mechanism. 
The results demonstrate that \methodname is highly robust to this hyperparameter. 
Performance for all models remains remarkably stable, regardless of whether early, middle, or final layers are selected. 

\begin{table}[t]
  \centering
  \caption{Evaluation of \methodname on the large-scale LLaVA-OV-72B~\cite{llavaov} model across the subtasks of the VidHalluc~\cite{vidhalluc}.}
  \vspace{-3mm}
  \label{tab:huge-model}
  \resizebox{0.475\textwidth}{!}{%
    \begin{tabular}{lccccc}
      \toprule
      \multirow{2}{*}{\textbf{Models}} 
      & \multicolumn{5}{c}{\textbf{VidHalluc}} \\
      \cmidrule(lr){2-6} 

      & BQA & MCQ & STH & TSH & AVG\\
      
      \midrule
      LLaVA-OV-72B~\cite{llavaov} & 76.80 & 90.79 & \textbf{42.74}  & 67.33 & 69.41\\
      \textbf{+\methodname (Ours)} & \textbf{78.46} & \textbf{90.93} & 40.43 & \textbf{76.00} & \cellcolor{blue!20}\textbf{71.46}\\

      \bottomrule
    \vspace{-2mm}
    \end{tabular}%
  }
\vspace{-4mm}
\end{table}

\vspace{-2mm}
\paragraph{Scalability to Large Scale VideoLLM.} 
To demonstrate the scalability and general applicability of our training-free framework, we applied \methodname to the large-scale LLaVA-OV-72B~\cite{llavaov} model. 
As shown in Tab.~\ref{tab:huge-model}, \methodname provides a consistent performance boost, improving the overall AVG score on VidHalluc~\cite{vidhalluc} by +2.05\%. 
The most significant improvement comes from the temporally-focused TSH subtask, which increases by +8.67\%. 
This result confirms that \methodname is a robust and scalable solution. 

\section{Conclusion}
\label{sec:conclusion}

In this paper, we introduce Self-Diagnostic Contrastive Decoding (\methodname), a training-free method that mitigates temporal alongside spatial hallucinations in VideoLLMs. To address temporal hallucination, we employ temporal homogenization to produce the ``temporally-hard'' negatives to expose the spurious temporal correlations and a self-diagnostic Mechanism that detects the hallucination tendencies of each token by measuring attention divergence across original video and negatives. With these introduced innovations, \methodname adaptively enhance spatial and temporal faithfulness for each output token. Extensive experiments demonstrate that \methodname substantially reduces hallucinations compared to existing methods, achieving state-of-the-art results on multiple video hallucination benchmarks while preserving general video understanding capabilities.
\clearpage
{
    \small
    \bibliographystyle{ieeenat_fullname}
    \bibliography{main}
}

\clearpage
\setcounter{page}{1}
\maketitlesupplementary

\section*{A. Detailed Experimental Settings}

\subsection*{A.1. Our hyperparameters ($\alpha$, $\beta$)}
To determine the optimal settings for \methodname, we performed a grid search over the hyperparameters $\alpha$ and $\beta$ for each benchmark. We explored the following configurations:
\[
(\alpha, \beta) \in \{(1.0, 0.33), (0.5, 0.25)\}
\]
This resulted in a total of 2 configurations evaluated for our proposed method.

\subsection*{A.2. Implementation Detail of Other Baselines}

\paragraph{Training-Free Baselines}
We compare our \methodname{} against two training-free baselines: TCD~\cite{eventhallusion} and DINO-HEAL~\cite{vidhalluc}. As the official implementations were not available at the time of writing, we re-implemented both methods strictly following the details provided in their respective papers. 
To ensure a fair comparison, we applied a similar grid search strategy to these baselines for each benchmark:

\vspace{0.5em} \noindent
TCD~\cite{eventhallusion}: We tuned the frame downsampling rate $r$ and the contrastive decoding parameters $(\alpha, \beta)$ over the following search space:
\[
    r \in \{2, 4\}, \quad (\alpha, \beta) \in \{(1.0, 0.1), (0.5, 0.5)\}
\]
This yields a total of $2 \times 2 = 4$ configurations.

\vspace{0.5em} \noindent
DINO-HEAL~\cite{vidhalluc}: We searched over two key components: normalization usage and DINO model variants.
\[
    \text{Normalization} \in \{\text{Enabled, Disabled}\}
\]
\[
    \text{DINO Variants} \in \{\text{With Registers, Without Registers}\}
\]
This results in $2 \times 2 = 4$ configurations.

\vspace{-0.5em} \noindent
\paragraph{Training-Based Baselines}
For training-based baselines: ArrowRL~\cite{arrowrl}, TPO~\cite{tpo}, and RRPO~\cite{rrpo}, we utilize the official pre-trained checkpoints provided by the respective authors. We strictly adhere to the model configurations specified within these checkpoints. All other implementation details and experimental settings are kept consistent with the original models to ensure the validity of the comparison.

\subsection*{A.3. Prompts among all Benchmarks}
We follow each benchmark's official provided prompt to implement our inference code. 

\section*{B. Additional Analysis}

\subsection*{B.1. Latency Report}
We report the inference latency to evaluate the computational cost of our proposed method. All measurements were conducted on a single NVIDIA H100 80GB GPU. 
Tab.~\ref{tab:latency} details the average inference time per sample (seconds/sample) on the VidHalluc~\cite{vidhalluc} benchmark.

\begin{table}[h]
    \centering
    \small
    \setlength{\tabcolsep}{8pt}
    \caption{\textbf{Per-sample inference latency} comparison on the VidHalluc~\cite{vidhalluc} benchmark. Results are reported in seconds.}
    \label{tab:latency}
    \begin{tabular}{lcccc}
    \toprule
      \textbf{Models} & \textbf{BQA} & \textbf{MCQ} & \textbf{STH} & \textbf{TSH}\\
    \midrule
    LLaVA-OV-7B~\cite{llavaov} & 0.80 & 0.84 & 0.89 & 1.04 \\
    +TCD~\cite{eventhallusion} & 0.94 & 1.00 & 1.06 & 1.20 \\
    +DINO-HEAL~\cite{vidhalluc} & 1.25 & 1.06 & 0.86 & 1.12 \\
    \textbf{+SEASON (Ours)} & 1.21 & 1.28 & 1.38 & 1.48 \\

    \midrule
    QWEN2.5-VL-7B~\cite{qwen25vl} & 0.77 & 0.87 & 0.92 & 1.12 \\
    +TCD~\cite{eventhallusion} & 0.90 & 1.03 & 1.07 & 1.24 \\
    +DINO-HEAL~\cite{vidhalluc} & 0.80 & 0.91 & 0.88 & 1.19 \\
    \textbf{+SEASON (Ours)} &  1.19 & 1.33 & 1.43 & 1.58\\
    
    \bottomrule
    \end{tabular}
    \vspace{-3mm}
\end{table}

\begin{figure}[h]
    \vspace{-2mm}
    \centering
    \includegraphics[width=0.8\linewidth]{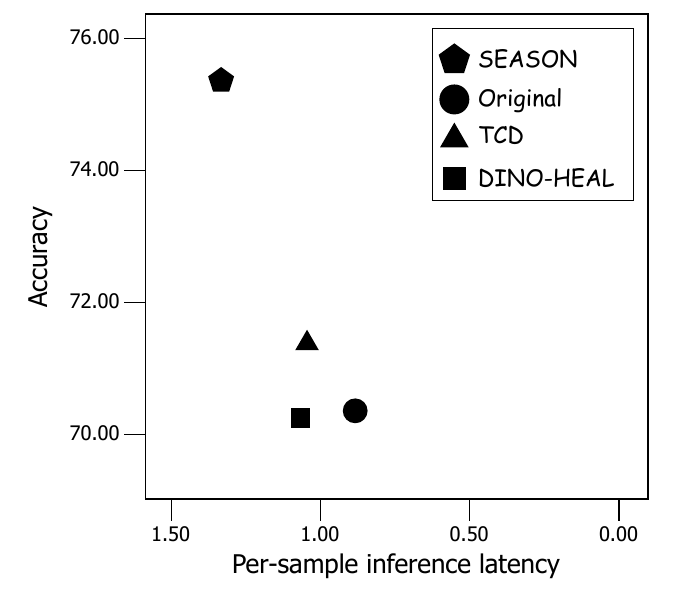} 
    \vspace{-2mm}
    \caption{Corresponding accuracy and latency of applying TCD~\cite{eventhallusion}, DINO-HEAL~\cite{vidhalluc}, and \methodname with LLaVA-OV-7B~\cite{llavaov} on VidHalluc~\cite{vidhalluc}.}
    \label{fig:latency_acc}
\end{figure}

\noindent
As expected, our method introduces moderate computational overhead compared to the base model due to additional operations (Sec.~\ref{sec:thv} and Sec.~\ref{sec:selfdiag}). However, as shown in Fig.~\ref{fig:latency_acc}, the latency remains within a reasonable range, while offering significant improvements in hallucination mitigation and preserving general video understanding (Tab.~\ref{tab:halbanchmark} and Tab.~\ref{tab:allbenchmarks}).

\subsection*{B.2. Hyperparameter Sensitivity ($\alpha$, $\beta$)}
\begin{figure}[h]
    \vspace{-1mm}
    \centering
    \includegraphics[width=0.7\linewidth]{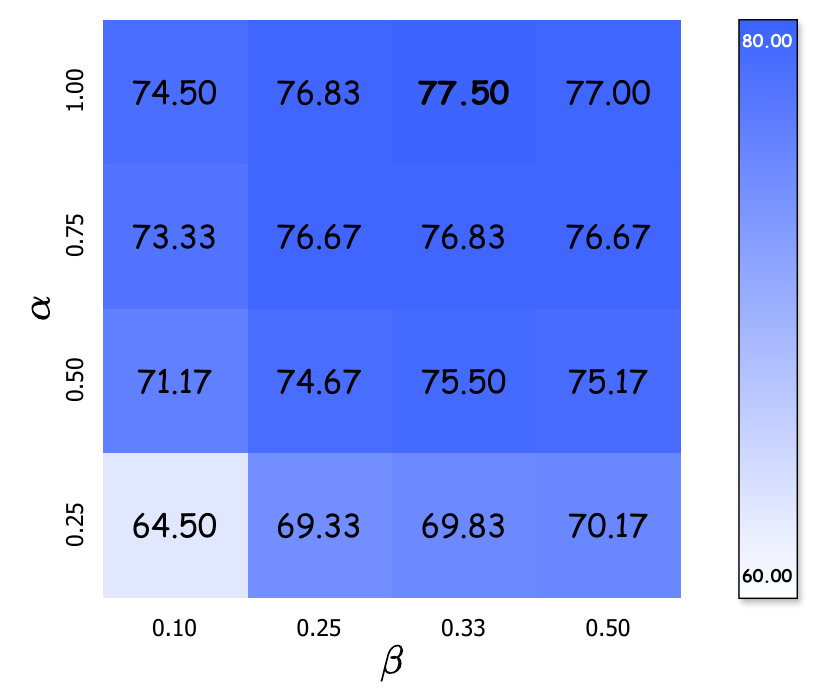} 
    \vspace{-2mm}
    \caption{Analysis of $\alpha$ and $\beta$. The values represent the Accuracy of LLaVA-OV-7B~\cite{llavaov} on the TSH subtask in VidHalluc~\cite{vidhalluc}.}
    \label{fig:heatmap}
\end{figure}

In Fig.~\ref{fig:heatmap}, we evaluated the performance on the TSH subtask in VidHalluc~\cite{vidhalluc} for the purpose of investigating the sensitivity of our method to the hyperparameters $\alpha$ and $\beta$.

\noindent
As observed in Fig.~\ref{fig:heatmap}, our method demonstrates consistent performance across a wide range of configurations. The performance on the TSH subtask in VidHalluc~\cite{vidhalluc} improves as $\alpha$ reaching 1.00 and $\beta$ reaching 0.33.

\subsection*{B.3. Temporal Homogenization Layers Ablation}
In Sec.~\ref{sec:thv}, we apply Temporal Homogenization at layers in the Model's Vision Encoder ($E_{\theta}$). Recall that at a given layer $l$ and frame $f_t$, the homogenized feature $h_{l,t}$ is defined as a linear combination of the frame feature from the corresponding global context $d_l$ and the pre-homogenization feature $h^{\prime}_{l,t}$ ($h_{0,t}$ are the patch embeddings of frame $f_t$):
\[
    h_{l,t} = (1 - \beta) h^{\prime}_{l,t} + \beta d_l, \, \text{where} \; h^{\prime}_{l,t} = \text{$E_{\theta}^{(l)}$}(h_{l-1,t}).
\]
By default, this operation is applied to all layers. 
In Tab.~\ref{tab:ablation_layers}, we vary the range of homogenization layers on LLaVA-OV-7B~\cite{llavaov}, in order to investigate the impact of layer selection in Model's Vision Encoder (e.g., Early, Middle, and Late).

\begin{table}[h]
    \centering
    \small
    \caption{Ablation study on the effect of applying Temporal Homogenization to different layers in Model's Vision Encoder ($E_{\theta}$).}
    \label{tab:ablation_layers}
    \begin{tabular}{lccc}
    \toprule
      \multirow{2}{*}{\textbf{Applied Layers}} 
      & \multicolumn{1}{c}{\textbf{VidHalluc}} 
      & \multicolumn{1}{c}{\textbf{VideoHallucer}} 
      & \multirow{2}{*}{\textbf{AVG}}\\
      \cmidrule(lr){2-2} \cmidrule(lr){3-3} 

      & TSH & TPH \\
    \midrule
    
    Early Layers  & 71.2 & 49.0 & 60.1 \\
    Middle Layers  & 74.2 & 53.0 & 63.6 \\
    Late Layers  & \underline{77.0} & \underline{54.5} & \underline{65.8} \\
    \textbf{All Layers (default)} & \textbf{77.7} & \textbf{55.5} & \textbf{66.6} \\
    
    \bottomrule
    \end{tabular}
\end{table}

\noindent
The results demonstrate that applying Temporal Homogenization to \textbf{All Layers} achieves the highest performance in temporal hallucination examination. Among the partial applications, Late Layers significantly outperform others. 

\section*{C. Additional Qualitative Evaluation}
\begin{figure*}[ht]
    \centering
    \includegraphics[width=0.95\textwidth]{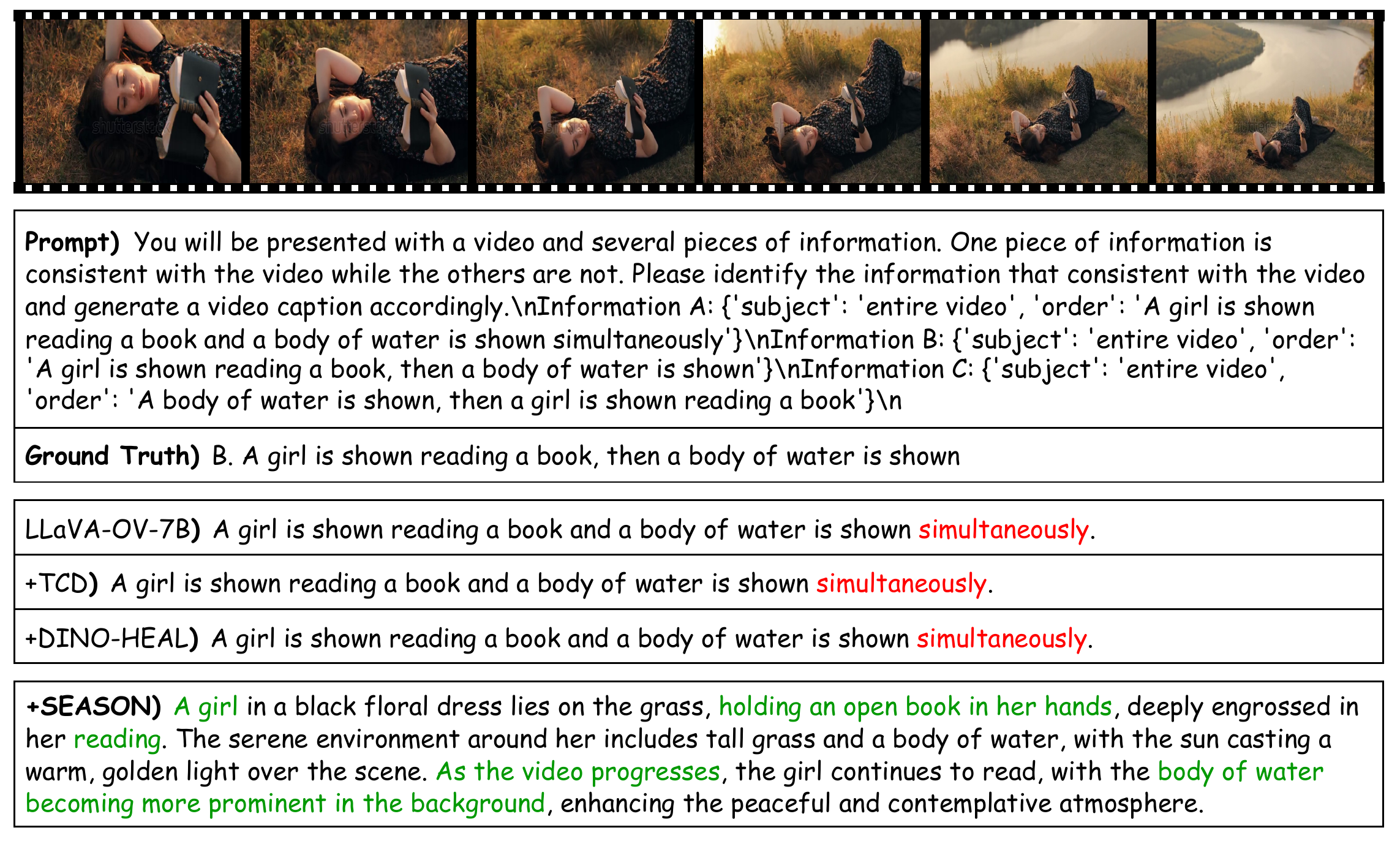} 
    \caption{Qualitative comparison of video captions predicted by LLaVA-OV-7B~\cite{llavaov} with TCD~\cite{eventhallusion}, DINO-HEAL~\cite{vidhalluc}, and \methodname on TempCompass~\cite{tempcompass}. Note that words highlighted
in \textcolor{OliveGreen}{\textbf{green}} indicate temporal faithfulness, while those in \textcolor{red}{\textbf{red}} indicate temporal hallucination.}
    \label{fig:supp_quali1}
\end{figure*}

\begin{figure*}[ht]
    \centering
    \includegraphics[width=0.95\textwidth]{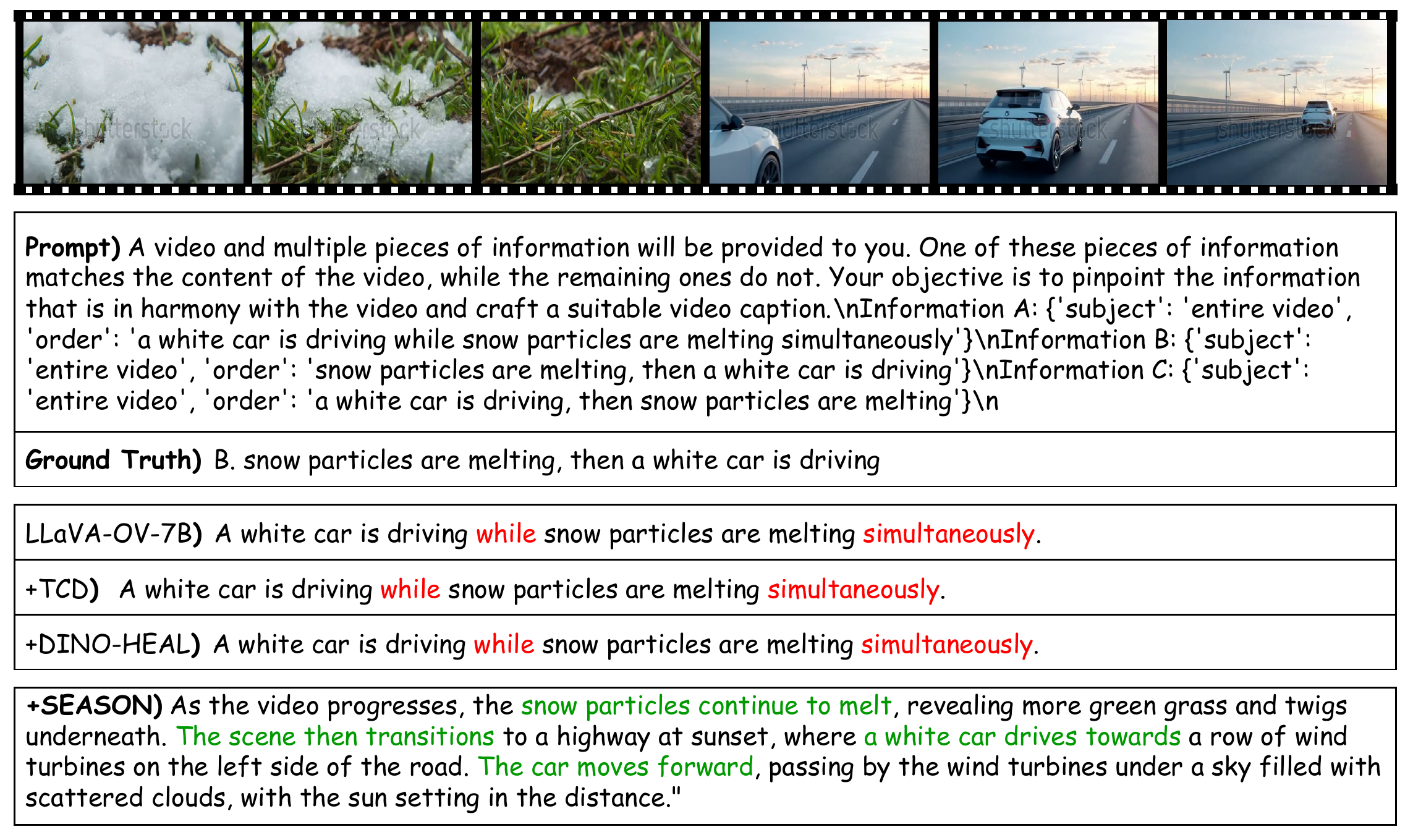} 
    \caption{Qualitative comparison of video captions predicted by LLaVA-OV-7B~\cite{llavaov} with TCD~\cite{eventhallusion}, DINO-HEAL~\cite{vidhalluc}, and \methodname on TempCompass~\cite{tempcompass}. Note that words highlighted
in \textcolor{OliveGreen}{\textbf{green}} indicate temporal faithfulness, while those in \textcolor{red}{\textbf{red}} indicate temporal hallucination.}
    \label{fig:supp_quali2}
\end{figure*}

\begin{figure*}[ht]
    \centering
    \includegraphics[width=0.95\textwidth]{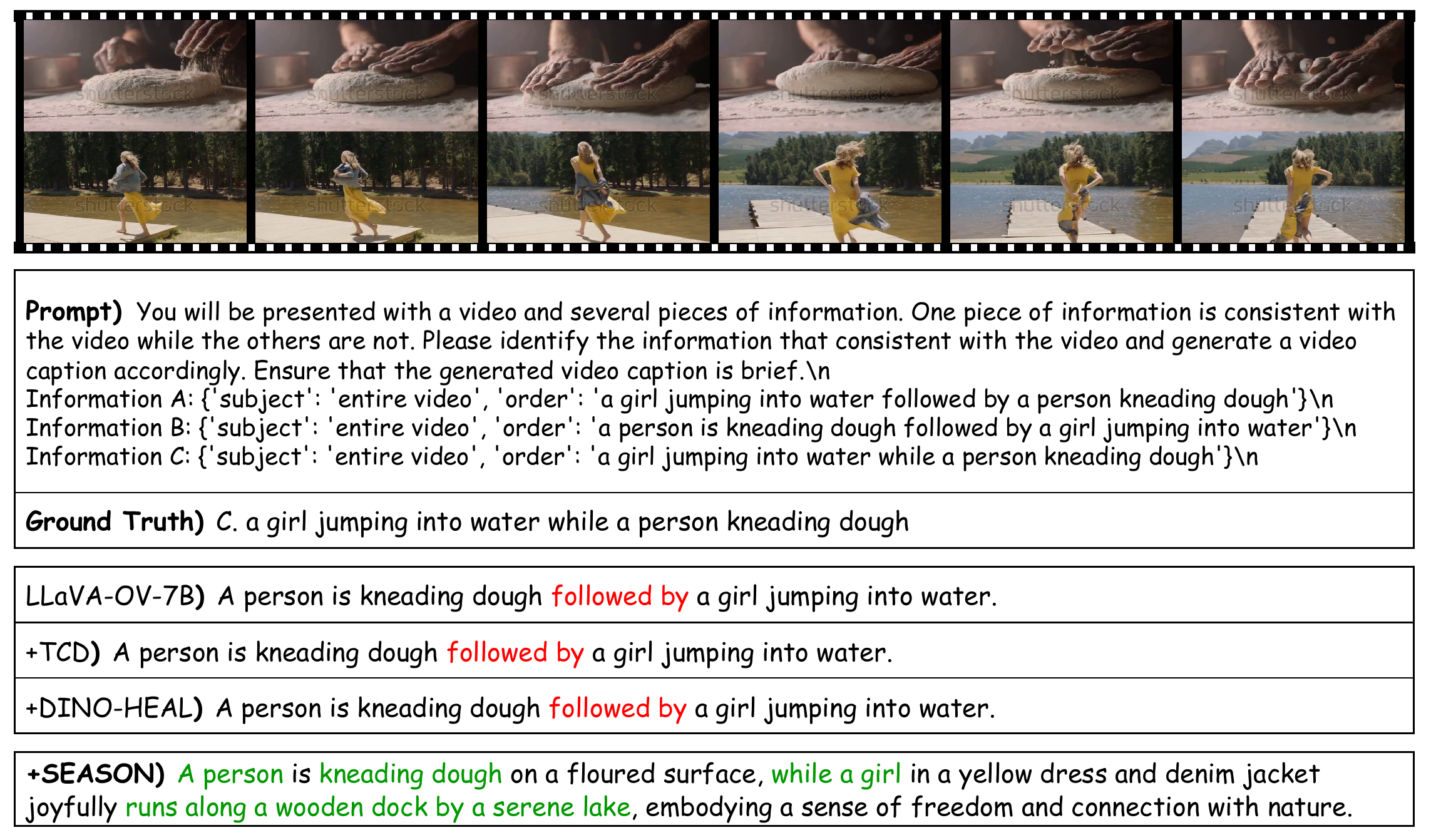} 
    \caption{Qualitative comparison of video captions predicted by LLaVA-OV-7B~\cite{llavaov} with TCD~\cite{eventhallusion}, DINO-HEAL~\cite{vidhalluc}, and \methodname on TempCompass~\cite{tempcompass}. Note that words highlighted
in \textcolor{OliveGreen}{\textbf{green}} indicate temporal faithfulness, while those in \textcolor{red}{\textbf{red}} indicate temporal hallucination.}
    \label{fig:supp_quali3}
\end{figure*}

\begin{figure*}[ht]
    \centering
    \includegraphics[width=0.95\textwidth]{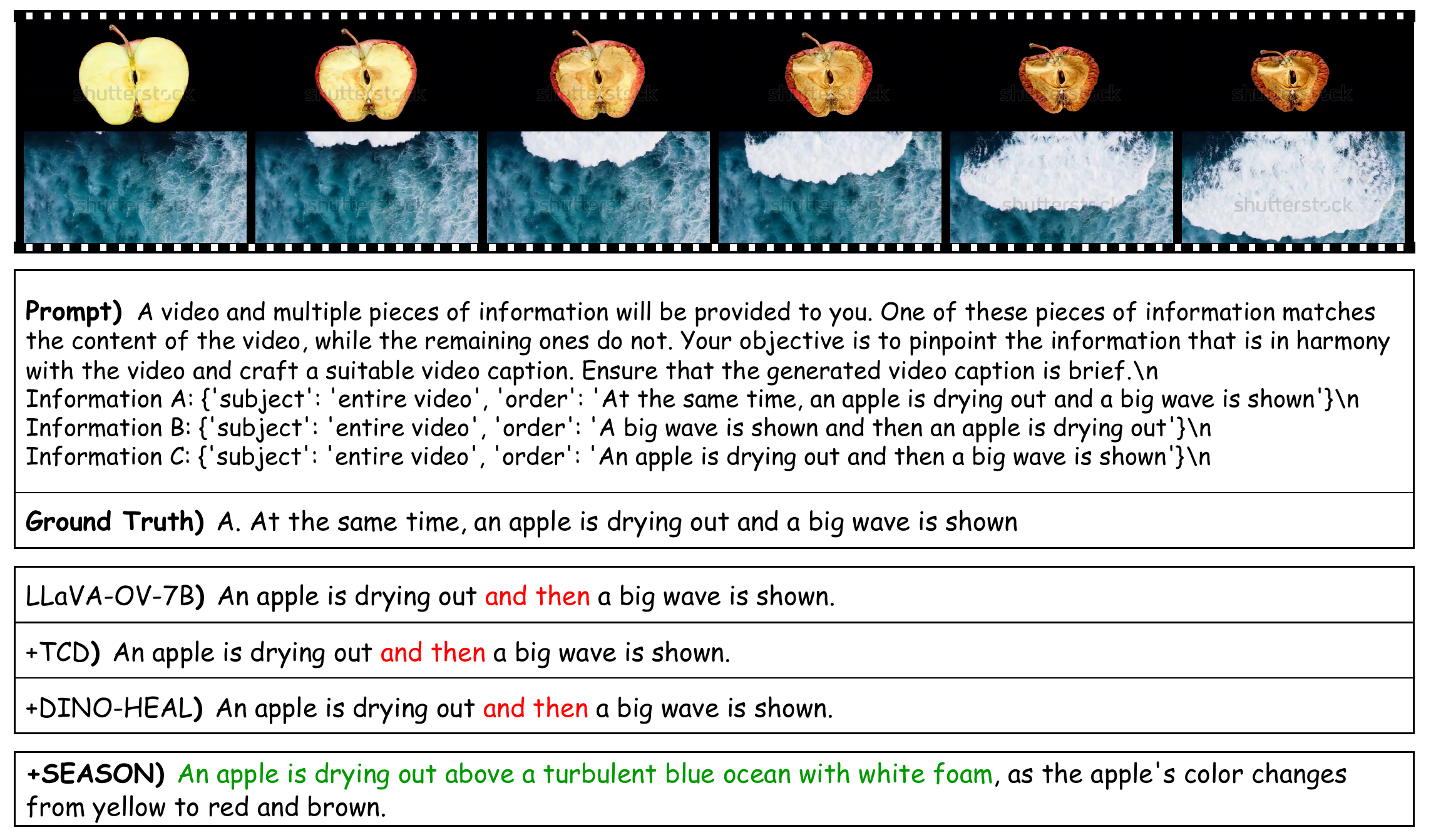} 
    \caption{Qualitative comparison of video captions predicted by LLaVA-OV-7B~\cite{llavaov} with TCD~\cite{eventhallusion}, DINO-HEAL~\cite{vidhalluc}, and \methodname on TempCompass~\cite{tempcompass}. Note that words highlighted
in \textcolor{OliveGreen}{\textbf{green}} indicate temporal faithfulness, while those in \textcolor{red}{\textbf{red}} indicate temporal hallucination.}
    \label{fig:supp_quali4}
\end{figure*}

\begin{figure*}[ht]
    \centering
    \includegraphics[width=0.95\textwidth]{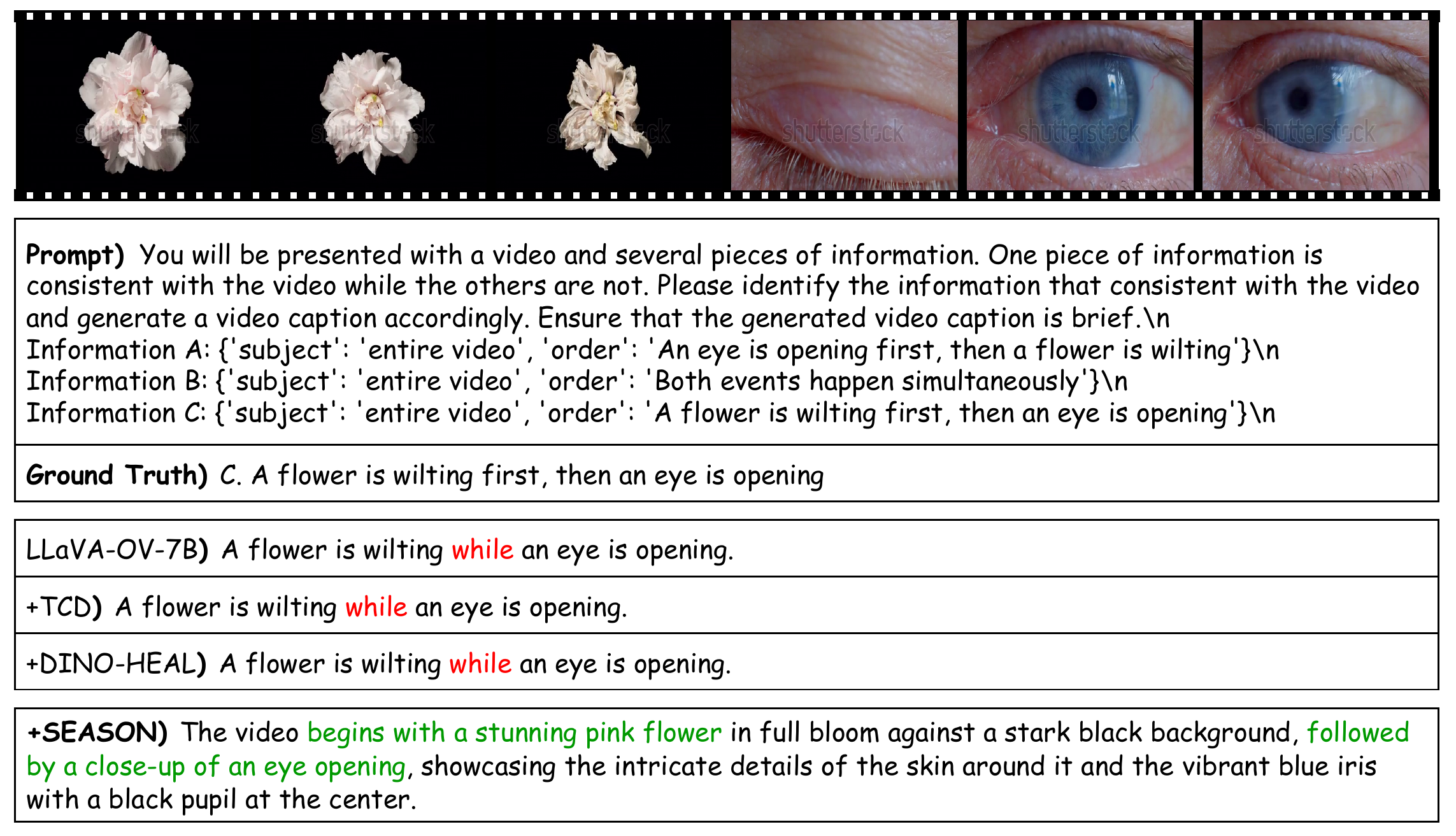} 
    \caption{Qualitative comparison of video captions predicted by LLaVA-OV-7B~\cite{llavaov} with TCD~\cite{eventhallusion}, DINO-HEAL~\cite{vidhalluc}, and \methodname on TempCompass~\cite{tempcompass}. Note that words highlighted
in \textcolor{OliveGreen}{\textbf{green}} indicate temporal faithfulness, while those in \textcolor{red}{\textbf{red}} indicate temporal hallucination.}
    \label{fig:supp_quali5}
\end{figure*}

\begin{figure*}[ht]
    \centering
    \includegraphics[width=0.95\textwidth]{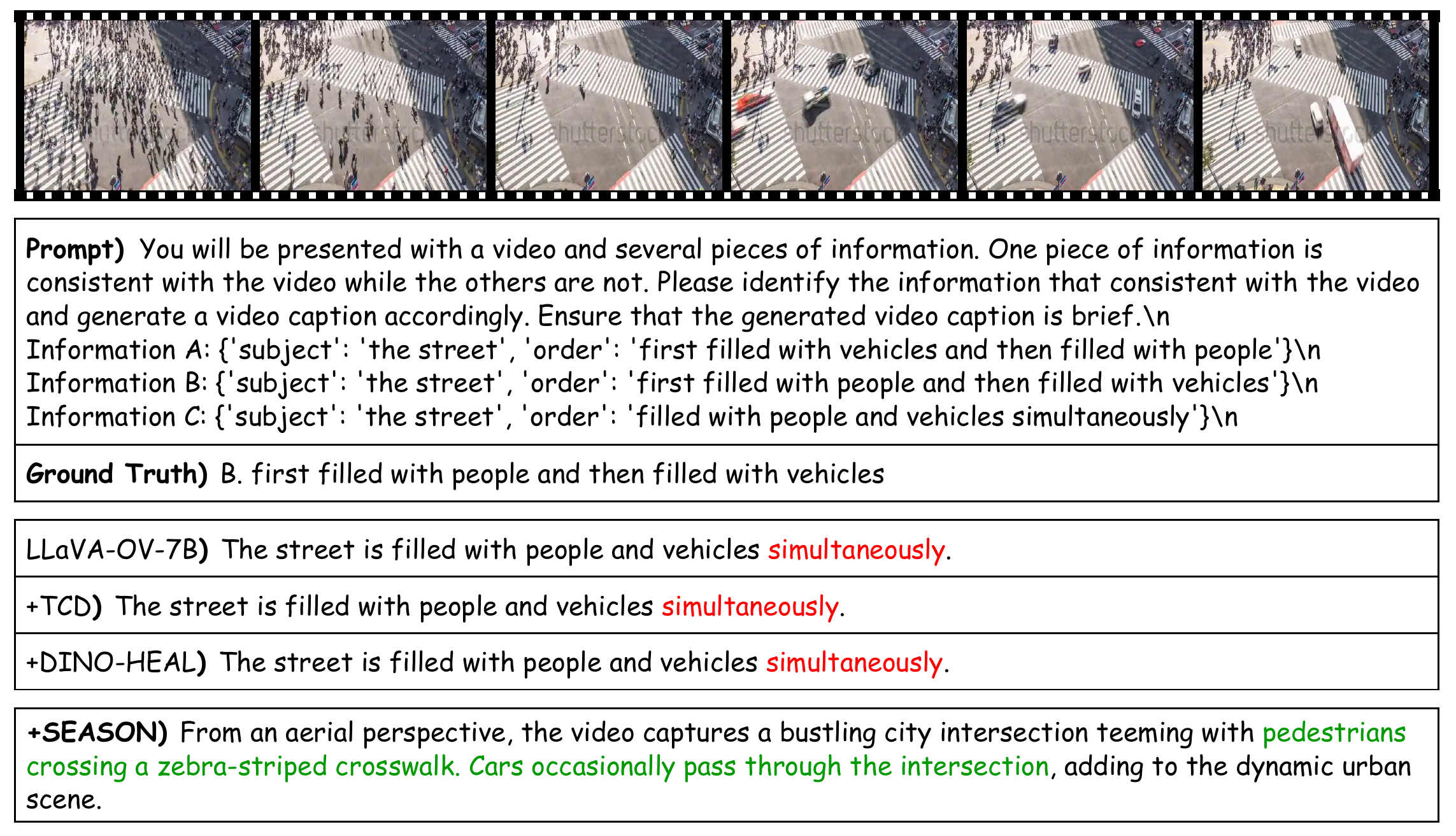} 
    \caption{Qualitative comparison of video captions predicted by LLaVA-OV-7B~\cite{llavaov} with TCD~\cite{eventhallusion}, DINO-HEAL~\cite{vidhalluc}, and \methodname on TempCompass~\cite{tempcompass}. Note that words highlighted
in \textcolor{OliveGreen}{\textbf{green}} indicate temporal faithfulness, while those in \textcolor{red}{\textbf{red}} indicate temporal hallucination.}
    \label{fig:supp_quali6}
\end{figure*}

\begin{figure*}[ht]
    \centering
    \includegraphics[width=0.95\textwidth]{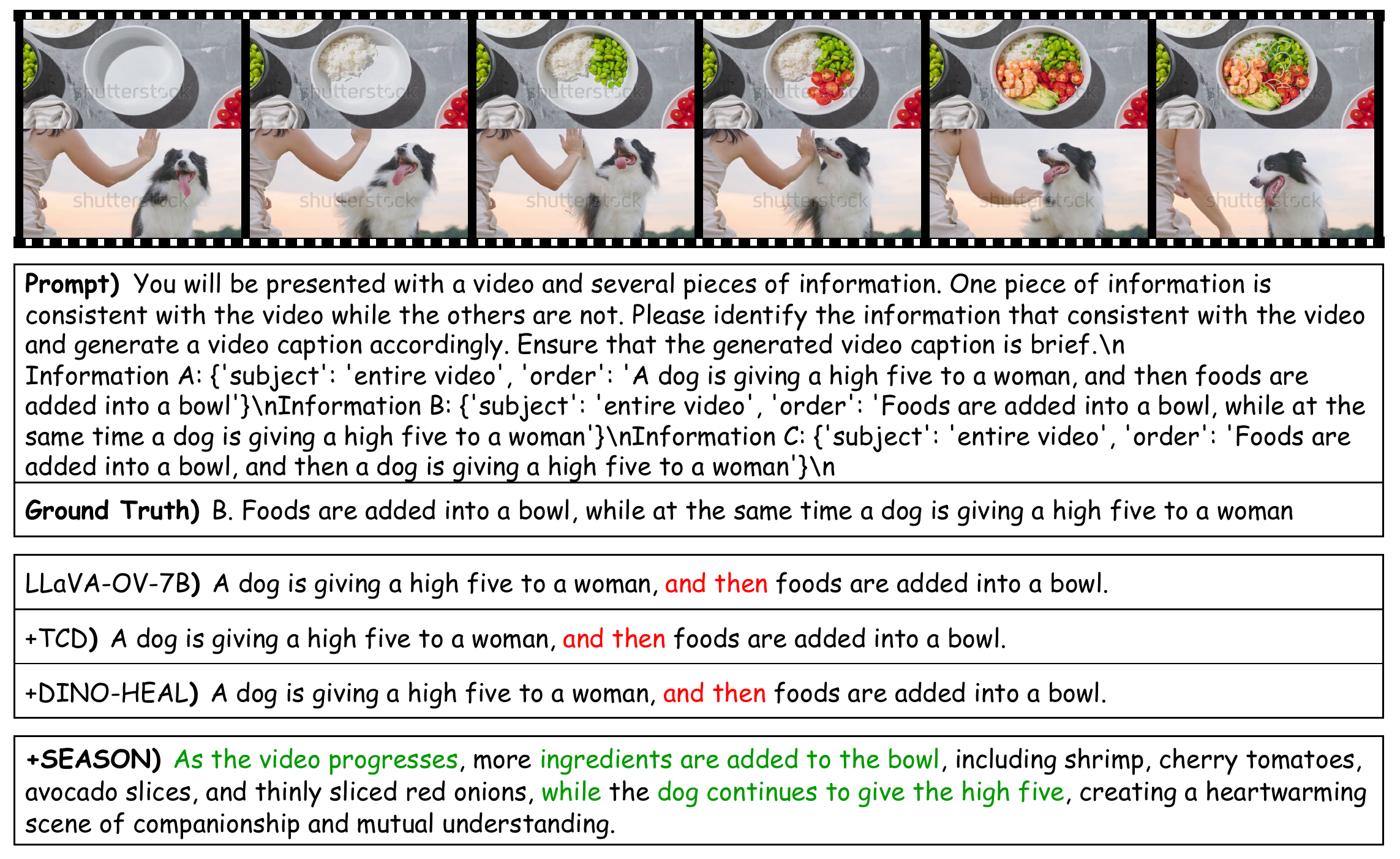} 
    \caption{Qualitative comparison of video captions predicted by LLaVA-OV-7B~\cite{llavaov} with TCD~\cite{eventhallusion}, DINO-HEAL~\cite{vidhalluc}, and \methodname on TempCompass~\cite{tempcompass}. Note that words highlighted
in \textcolor{OliveGreen}{\textbf{green}} indicate temporal faithfulness, while those in \textcolor{red}{\textbf{red}} indicate temporal hallucination.}
    \label{fig:supp_quali7}
\end{figure*}

In Figs.~\ref{fig:supp_quali1} to~\ref{fig:supp_quali7}, we present additional qualitative results of applying TCD~\cite{eventhallusion}, DINO-HEAL~\cite{vidhalluc}, and \methodname with LLaVA-OV-7B~\cite{llavaov} on TempCompass~\cite{tempcompass}. \methodname exhibits temporal faithfulness within its generated captions, which demonstrate the effectiveness of \methodname in mitigating temporal hallucinations.

\end{document}